\documentclass{article} 
\usepackage{iclr2025_conference,times}

\usepackage{amsmath,amsfonts,bm}

\def\eqref#1{equation~\ref{#1}}

\def\1{\bm{1}}

\DeclareMathAlphabet{\mathsfit}{\encodingdefault}{\sfdefault}{m}{sl}
\SetMathAlphabet{\mathsfit}{bold}{\encodingdefault}{\sfdefault}{bx}{n}

\usepackage{hyperref}
\usepackage{url}
\usepackage{amsmath}
\usepackage{amssymb}
\usepackage{mathtools}
\usepackage{amsthm}
\usepackage[english]{babel}
\usepackage{blindtext}
\usepackage[hypcap=false]{subcaption}
\usepackage[hypcap=false]{caption}
\usepackage{multirow}
\usepackage{array}
\usepackage[11pt]{moresize}
\usepackage{xspace}
\usepackage{arydshln}
\usepackage{graphicx}
\usepackage{amsfonts}
\usepackage{booktabs}
\usepackage{gensymb}
\usepackage{color}
\usepackage{natbib}
\usepackage{soul}
\usepackage{makecell}
\usepackage{bm}
\usepackage{algorithm}
\usepackage[noend]{algpseudocode}
\usepackage{wrapfig}
\theoremstyle{plain}
\newtheorem{theorem}{Theorem}[section]

\newtheorem{lemma}[theorem]{Lemma}

\theoremstyle{definition}

\theoremstyle{remark}

\newenvironment{tightlist}{
\begin{list}{$\bullet$}{
    \setlength{\topsep}{.1em}
    \setlength{\partopsep}{0in}
    \setlength{\parskip}{0in}
    \setlength{\itemsep}{0in}
    \setlength{\parsep}{0in}
    \setlength{\leftmargin}{1em}
    \setlength{\rightmargin}{0in}
    \setlength{\itemindent}{0in}
}}
{\end{list}}

\title{PH-Dropout: Practical Epistemic Uncertainty Quantification for View Synthesis}

\author{Chuanhao Sun$^{1}$, Thanos Triantafyllou$^{1}$, Anthos Makris$^{1}$, Maja Drmač$^{1}$, Kai Xu$^{2}$, \\ \textbf{Luo Mai}$^{1}$, \textbf{Mahesh K. Marina}$^{1}$\\
$^{1}$ School of Informatics, The University of Edinburgh, Edinburgh, UK\\
$^{2}$ MIT-IBM Watson AI Lab, Cambridge, MA, US\\
\texttt{\{chuanhao.sun,mahesh\}@ed.ac.uk} 
}

\newcommand{\method}{\textsc{PH-Dropout}\xspace}

\iclrfinalcopy 
\begin{document}

\maketitle

\begin{abstract}

View synthesis using Neural Radiance Fields (NeRF) and Gaussian Splatting (GS) has demonstrated impressive fidelity in rendering real-world scenarios. However, practical methods for accurate and efficient epistemic Uncertainty Quantification (UQ) in view synthesis are lacking. Existing approaches for NeRF either introduce significant computational overhead (e.g., ``10x increase in training time" or ``10x repeated training") or are limited to specific uncertainty conditions or models. Notably, GS models lack any systematic approach for comprehensive epistemic UQ. This capability is crucial for improving the robustness and scalability of neural view synthesis, enabling active model updates, error estimation, and scalable ensemble modeling based on uncertainty. In this paper, we revisit NeRF and GS-based methods from a function approximation perspective, identifying key differences and connections in 3D representation learning. Building on these insights, we introduce \method (\textbf{P}ost \textbf{h}oc Dropout), the first real-time and accurate method for epistemic uncertainty estimation that operates directly on pre-trained NeRF and GS models. Extensive evaluations validate our theoretical findings and demonstrate the effectiveness of \method.
\par \textbf{Code available at}: \url{https://github.com/thanostriantafyllou3/ph-dropout}
\end{abstract}

\section{Introduction}

Emerging approaches in view synthesis, such as Neural Radiance Fields (NeRF)~\citep{mildenhall2021nerf} and Gaussian Splatting (GS)~\citep{kerbl20233d}, have demonstrated remarkable advancements in rendering quality and efficiency. These methods transcend synthetic datasets to embrace real-world, unconstrained scenarios, setting new standards for generating highly realistic 3D scenes that are nearly indistinguishable from reality. However, achieving high-quality results requires a large number of known views for training, ensuring that the model is exposed to multiple perspectives near any arbitrary target viewpoint. Previous works~\citep{goli2024bayes, sunderhauf2023density, hu2024cg} have highlighted that training view synthesis models from a discrete set of multiview images is fraught with uncertainty. Even under ideal experimental conditions, occlusions and missing views inherently limit the epistemic knowledge that the model can acquire about the scene.


Studying epistemic uncertainty in view synthesis is crucial for understanding the limitations of algorithms, identifying gaps in model performance, improving the reliability of predictions, and ensuring effective generalization to unseen data. This investigation is pivotal for advancing the robustness and accuracy of view synthesis methods, yet research in this area remains limited. Broadly, related work can be classified into two categories: (1) Direct estimation of overall epistemic uncertainty, as seen in methods such as S-NeRF~\citep{shen2021stochastic}, CF-NeRF~\citep{shen2022conditional}, and NeRF OTG~\citep{ren2024nerf}, which apply deep ensemble techniques to NeRF with significant computational overhead. (2) Investigation of specific factors contributing to epistemic uncertainty, such as Bayes Rays~\citep{goli2024bayes} and CG-SLAM~\citep{hu2024cg}, which focus primarily on spatial (depth) uncertainty caused by a lack of views. In fact, epistemic uncertainty can be caused by a range of factors beyond mere lack of training data, including inadequate feature representation and model misspecification. Traditional methods for estimating epistemic uncertainty, such as Monte-Carlo dropout~\citep{gal2016dropout} and random initialization~\citep{lee2015m,lakshminarayanan2017simple}, prove impractical for view synthesis due to conflicts with the training paradigm or the prohibitively high computational demands, requiring hours even for simple scenarios.

Another limitation is efficiency. Ensemble-based approaches can require hours of additional training for a single bounded object. Furthermore, there is a noticeable gap between GS and NeRF models, as existing methods lack versatility and are typically applicable to only one of these frameworks. However, we observe that NeRF-based and GS-based solutions excel in different scenarios. A versatile method integrating both approaches could enable the development of a new rendering technique that combines their respective strengths.

Driven by the goal of developing a more efficient, accurate, and versatile approach for estimating epistemic uncertainty in view synthesis, we revisit current view synthesis methods through the lens of function learning and approximation theory. We make the following key observations:
\begin{tightlist}
    \item Existing view synthesis models often exhibit \textbf{substantial parameter redundancy}; specifically, their performance on the training set remains unaffected by the application of appropriate dropout.
    \item However, while no observable impact is seen on training views, significant performance variance is evident on test views.
\end{tightlist}
These insights lead us to propose a relatively underexplored post hoc epistemic uncertainty estimation method: (1) insert dropout on trained fully connected layers (NeRF) or splats (GS); (2) increase the dropout ratio as long as the performance on the training set remains unaffected; (3) measure the variation after dropout on the testing set as a quantification of epistemic uncertainty (UQ).

In summary, our contributions are:
\begin{tightlist}
    \item We propose \method, the first approach, to our knowledge, that quantifies epistemic uncertainty for view synthesis in real-time, orders of magnitude faster, without additional training, and applicable to both NeRF and GS methods.
    \item We conduct comprehensive experiments with \method on NeRF-based and GS-based methods, providing the first in-depth comparison of their features beyond training speed and fidelity.
    \item We evaluate the effectiveness of \method through several downstream use cases, including active learning and model ensembling, where it demonstrates promising performance in supporting these applications.
\end{tightlist}

\section{Background}
\subsection{Epistemic UQ with Specialized Training Approach}
We explore general methods for epistemic UQ and explain why they are unsuitable for typical view synthesis tasks, highlighting the need for a novel approach. These traditional methods impose significant computational overhead and impose strict limitations on model selection.

\textit{Random initialization}~\citep{lee2015m, lakshminarayanan2017simple}: is one technique used to measure epistemic uncertainty, particularly in the context of deep learning. However, it is not the most comprehensive method for capturing all aspects of uncertainty for following reason.
\begin{tightlist}
\item \textbf{Training overhead}. Repeatedly retraining the model can be impractical, especially when the training process is slow or computationally expensive.
\item \textbf{Random initialization cannot address limitations inherent in the model architecture itself}. If the model is not capable of learning the true underlying distribution, then random initialization won't reveal this inadequacy clearly. 
\end{tightlist}

\textit{Monte Carlo Dropout}~\citep{gal2016dropout}: Applying dropout at test time and averaging predictions can be a cheaper way to estimate uncertainty without needing to train multiple models. NeRF's tendency to overfit arises from its high model capacity, the limited and specific nature of its training data, and its per-scene training approach. However, dropout is a method to prevent overfitting, which prevents NeRF to memorize the scene in training set. Empirically, we find that training with dropout in conventional NeRF places negative effect on performance, and hence MC dropout is not suitable for NeRF. Also GS models do not have a neural network, and hence hard to implement BNN based methods. Similar methods can be find in MC-batchnorm~\citep{teye2018bayesian}, where a deterministic network trained with batch normalization, which is also maintained during testing for UQ. Detailed discussion on applying MC-dropout on NeRF is enclosed in Appendix~\ref{ap:mc}

\textit{Deep Ensemble Methods}~\citep{lakshminarayanan2017simple}: Training multiple models (not just with different initializations but potentially with different architectures or hyperparameters) and aggregating their predictions can provide a more robust estimate of uncertainty. S-NeRF~\citep{shen2021stochastic}, CF-NeRF~\citep{shen2022conditional}, and 3D Uncertainty Field~\citep{shen2024estimating} are based on this concept but with expensive training overhead and poor performance.

\subsection{Post hoc epistemic UQ}
Post hoc epistemic UQ refers to techniques for assessing a trained model's epistemic uncertainty without altering its original training process. The most frequently used frameworks for modeling uncertainty of neural networks are often not agnostic to the network architecture and task, and also require modifications in the optimization and training processes, including MC-dropout and deep ensembles. This is why post hoc UQ is desirable -- it can be applied to already
trained architectures. 

We distinguish our study with the epistemic UQ in well pretrained models~\citep{wang2024epistemic, NEURIPS2023_3e0b9620}. By well pretrained, the basic assumption of this method is the model is trained with sufficient in-distribution training data $\mathcal{D}$, so that
$\forall x\in \mathcal{X}\rightarrow p(x\notin\mathcal{D})<\epsilon$, where $\mathcal{X}$ is the set of potential inputs, $\epsilon$ is a small positive number. However, typical view synthesis tasks often face missing training views, making such methods unsuitable. Therefore, we exclude them as baselines in this paper.

Bayes Rays~\citep{goli2024bayes}, based on Laplace approximation~\citep{ritter2018scalable}, meets the post hoc requirements for practical application, requiring only a few additional training epochs. However, it solely models spatial uncertainty (depth prediction error) in NeRF and is not applicable to GS-based models. Previous work~\citep{ledda2023dropout} has also shown that in ad network based on fully connected layers, inject dropout during the inference time can achieve similar effect as MC-dropout~\citep{gal2016dropout} after calibration. However, it is non-trivial to apply this method to view synthesis model. NeRF does not use dropout during the training, so it is hard to justify the approximation of MC-dropout. GS model even does not have a typical neural network. 
\section{\method for Epistemic Uncertainty Estimation}
In this section, we first introduce the proposed algorithm in \S\ref{subsec:phd}, then provide the conditions to ensure the effectiveness of the proposed method.
\subsection{\method}\label{subsec:phd}
Here we introduce the proposed algorithm \method. The process of \method is illustrated in the following pseudo-code,
\begin{algorithm}
\caption{\method}\label{alg:phd}
\begin{algorithmic}[1] 
\Require Trained model: $F(\cdot;\theta)$, threshold $\epsilon$, step length $\Delta_r$, sampling number $N$
\Ensure $\mathbb{E}_x(|F(x;\theta)-F(x;\mathrm{PHD}(\theta,r))|)<\epsilon$
\State $r \gets \Delta_r$ 
\Comment{Initialize dropout ratio.} 
\State $D(\theta,r)\gets M\cdot\theta$, $M_{ij} \in \{0, 1\}, \forall i, j$ and $r=\frac{\sum M_{ij}}{|\theta|}$\Comment{$M:\{M_{ij}\}$ is the binary dropout mask}
\State $\mathbb{E}_x(|F(x;\theta)-F(x;\mathrm{PHD}(\theta,r))|)\gets\frac{1}{N}\sum^N_i\sum_{x\in\mathcal{X}}\frac{|F(x;\theta)-F(x;D_i(\theta,r))|}{|\mathcal{X}|}$ \Comment{$D_i(\theta,r)\sim\mathrm{PHD}(\theta,r)$}
\While{$\mathbb{E}_x(|F(x;\theta)-F(x;\mathrm{PHD}(\theta,r))|)<\epsilon$}
\State $r\gets r+\Delta_r$ \Comment{Increase dropout ratio}
\EndWhile
\State $r_{\text{drop}}\gets r-\Delta_r$\Comment{Select the maximal $r$ that meets requirement}
\If{$r_{\text{drop}} = 0$}\Comment{$r_{\text{drop}}=0\rightarrow$Wrong configuration}
\State Raise Error \Comment{The model is not properly trained}
\Else
\State $\zeta(x), \gets \mathrm{std}(F(x;\mathrm{PHD}(\theta,r_{\text{drop}})))$ 
\Comment{Per pixel and channel UQ $\zeta$ of input $x$ based on std}
\State $\overline{\sigma_{\text{max}}}\gets G(\zeta(x))$
\Comment{$G(\cdot)$ represents the processing in \S\ref{ap:sigma}}
\EndIf
\end{algorithmic}
\end{algorithm}
where $F(x; \theta)$ is the trained rendering function (NeRF or GS) with parameters $\theta$, $\mathbb{E}(\cdot)$ denotes the expectation, $\mathrm{PHD}(\cdot)$ refers to repeating stochastic inferences using independent dropout masks $M$, and $D(\cdot)$ is a sample of $\mathrm{PHD}(\cdot)$. It includes a heuristic solution of $\arg\max_r\mathbb{E}_x(|F(x;\theta)-F(x;\mathrm{PHD}(\theta,r))|)<\epsilon$. If dropout ratio $r_{\text{drop}}=0$ after interaction, then this indicates the model is configured with less parameter than need (see Theorem~\ref{theo:must_redund}). $r_{\text{drop}}$ is a measurement of parameter redundancy.

To further quantify the overall uncertainty $\zeta(x)$ of a model, we introduce $\overline{\sigma_{\text{max}}}$ as a metric, which considers the max std across height, width, and RGB channels for each image, averaged over the whole rendered image set. See detailed definition in \S\ref{ap:sigma}. By default, we use $\overline{\sigma_{\text{max}}}$ to represent the overall uncertainty of a trained model on a given scene.

The differences to inject dropout in \cite{ledda2023dropout} are as follows,
\begin{tightlist}
\item The dropout is applied with the condition that the model must still perfectly fit the training set afterward.
\item Unlike a standard dropout layer, the dropout mask in \method directly sets the weights (or splats in GS) to zero without scaling up amplitude of the rest weights.
\item In NeRF-based methods, we apply dropout (i.e., a binary mask) to one of the middle layers, typically after the first fully connected layer, to selectively remove components from the render function, enhancing control over the process.
\end{tightlist}

\subsection{Conditions of Using \method: Features in View Synthesis}
The following phenomenons are integral to our reasoning and will be validated through experiments.

\textbf{Phenomenon 1}: The rendering function is not stochastic since we try to render a static object. 

\textbf{Phenomenon 2}: After dropout, as long as there is no change on training set performance, the expectation performance on evaluation set only has negligible change as well (see \S\ref{sec:eval}).

Besides empirical observations, we also notice a common features across all NeRF-based and GS-based method: there must be redundancy in model parameters $\theta$, which is explained in detail in Theorem~\ref{theo:must_redund}.

\begin{theorem}\label{theo:must_redund}
As long as the model is properly trained with overfitting ($\mathcal{L}(x)\rightarrow 0$ on training set), there must be significant redundancy in NeRF and GS model, \emph{i.e.}, 
\[
\exists~0\ll r<1 \rightarrow \forall x \in \mathcal{D}_{\text{train}},~ |F(x;\theta) - F(x;D(\theta, r))| < \epsilon
\]
\end{theorem}

\begin{proof}
(\textbf{Sketch}) The rendering function is neither purely continuous nor purely discrete, but rather a combination of both. Achieving fine convergence using exclusively continuous methods (e.g., NeRF) or discrete methods (e.g., 3DGS) requires an infinite number of Fourier components (as in NeRF with positional encoding) or splats (as in splatting-based methods like 3DGS) to approximate functions with varying degrees of continuity.  

\textit{NeRF: Continuous function $\rightarrow$ discrete representation}. For many natural signals~\citep{oppenheim1997signals}, 
 the amplitude of Fourier coefficients $c_n$ decay rapidly as 
$|n|$ increases, especially when signal is ``continuously differentiable''. This means that the lower-frequency components (those with smaller $|n|$) contain most of the signal's power, while the higher-frequency components (larger 
$|n|$) contribute very little to the total power. To overfit a function with nearly discrete pattern, many low power and high frequency components are introduced.

\textit{GS: Discrete splats $\rightarrow$ continuous representation}. Similar to the Fourier transform, where fine spatial details (higher frequency components) generally have lower power, the training of GS models follows a similar pattern. A few large splats are used to capture the broader, background features, while numerous smaller splats are introduced to capture finer details and subtle variations in the rendering function.

As a result, in both NeRF and GS, to overfit the details of rendering function, most of the components have very low power and therefore are robust to dropout.
\end{proof}
Theorem~\ref{theo:must_redund} also highlights that NeRF-based and GS-based methods exhibit varying levels of redundancy at different regions of an object or scene. Both experimental results and theoretical analysis later demonstrate that this variation in redundancy patterns contributes to distinct robustness and reliability when capturing specific features. By leveraging an ensemble approach that selects the components with the lowest epistemic uncertainty, we achieve a significant improvement in view synthesis fidelity.

Empirical results show that all well-converged view synthesis models exhibit $20–30\%$ redundancy, varying across methods. Removing this redundancy causes significant convergence issues during training. Thus, we conclude that \method \textbf{does not require an additional increase in parameters, as this redundancy is by default required for proper convergence.}

\section{Effectiveness of Estimation with \method}
\subsection{Effectiveness Analysis of \method in NeRF}
Here we discuss NeRF based methods with encoding that can reflect the spacial proximity faithfully, including Positional Encoding (PE)~\citep{tancik2020fourier}, Sinusoidal PE (SPE)~\citep{DBLP:conf/icml/SunYXMSCM24}, etc. We discuss hash encoding-based methods separately in Section \S\ref{subsec:collision} due to the impact of their unique probabilistic operations.
\begin{lemma}\label{theo:not_sparse}
If two models $a$ and $b$ with same structural and number of parameter, have similar distribution of parameter, \textbf{i.e.}, $D_{\mathbf{KL}}(\theta_a, \theta_b)<\epsilon$, and they both converge on the same dataset $\mathcal{D}$, they can be obtained via random initialization with the same setup with $a$ or $b$. Meanwhile, the probability density to obtain model $a$ and $b$ will be close, \emph{i.e.}, if $p(a)$ is significant, then $p(b)$ should be significant as well.
\end{lemma}

\begin{proof}
(\textbf{Sketch}). This can be proven based on the continuousity of the space of model parameter. The random initialization of weights of MLP (\emph{e.g.,} in NeRF) from continuous distributions, which influences the training dynamics~\cite{glorot2010understanding}. These continuous distributions provide the network's starting point for training, and as training proceeds, the weights and biases are updated continuously by optimization algorithms (\emph{e.g.}, stochastic gradient descent). These updates are applied to real-valued weights, thus ensuring that the neural network's parameters remain in the continuous space $\mathbb{R}^d$, where $d$ is the number of parameters~\cite{raghu2017expressive}. Because the function space $\mathbb{R}^d$ is continuous, if the two models have small KL partition, \emph{i.e.}, $D_{\mathbf{KL}}(\theta_a, \theta_b)<\epsilon$, then  $p(a)\approx p(b)$ in random initialisation because of the continuousity.
\end{proof} 

With Lemma~\ref{theo:not_sparse}, we establish a connection between random initialization~\citep{lakshminarayanan2017simple, lee2015m} and \method, where each ensemble in \method should exhibit significant probability density, assuming the trained model (w/o dropout) maintains substantial probability density within the function space. This forms the following Theorem.
\begin{theorem}\label{theo:effectiveness}
The variance after \method represents a biased epistemic uncertainty estimation.
\end{theorem}
\begin{proof}
(\textbf{Sketch}) With \method, the KL divergence of the remaining parameters from the original model is small. For instance, in NeRF, dropout is applied to only one hidden layer, leaving most parameters unchanged. Similarly, in the GS model, most splats have low power (see Proof of Theorem~\ref{theo:must_redund}), and setting them to zero minimally impacts the overall power distribution. 

We can further assume that the trained model is not an outlier and that the probability density of obtaining 
$F$ after random initialization is significant. Consequently, models after dropout represent a biased subset of the ensemble, and their variation captures epistemic uncertainty (see detailed discussion in Appendix~\ref{ap:effective}, Theorem~\ref{thm:ensemble}), as the probability density of these functions is non-negligible, consistent with Lemma~\ref{theo:not_sparse}.
\end{proof}

We have demonstrated that \method is an effective approach for producing ensembles in NeRF, with the ensemble variation reflecting a biased estimation of epistemic uncertainty. Next, we extend this reasoning to the GS model.

\subsection{Epistemic Uncertainty Estimation in Gaussian Splats}
\begin{theorem}\label{theo:gs_continual}
During the training of 3DGS and 2DGS models following the scheme in \citep{DBLP:journals/tog/KerblKLD23}, the state of the splats in later training phase can be taken as changing in a continuous space, \emph{i.e.}, the probability distribution density function $\mathcal{P}$ of model parameters $\theta$ is continuous.
\end{theorem}

\begin{proof}
According to \citep{kheradmand20243d}, we know the typical splats updating scheme in \citep{DBLP:journals/tog/KerblKLD23} can be approximated by a Stochastic Gradient Langevin Dynamics \citep{brosse2018promises}, with a noise term missing. Standard Gaussian Splatting optimization could be understood as having Gaussians that are sampled from a likelihood distribution that is tied to the rendering quality. Suppose $\mathcal{P}$ is the data-dependent probability density function of models, it will have a form $\mathcal{P}\propto\exp{(-\mathcal{L})}$, where $\mathcal{L}$ is the loss function during training. Because the loss function $\mathcal{L}$ is continuous, the density function $\mathcal{P}$ should be continuous as well.
\end{proof}

Because of the continuousity of GS updating dynamics in Theorem~\ref{theo:gs_continual}, we can extend the Lemma~\ref{theo:not_sparse} to typical GS models. Similar to dropout in the fully connected layers, we directly dropout the Gaussians, as they are the ``weights'' to optimize during the training.

\subsection{\method is unable to handle input hash collision}\label{subsec:collision}
NeRF may use hash encoding (HE) to process the input $x$. When collision happens, we have $x_i\neq x_j$ and $h(x_i) \approx h(x_j)$, where $h(\cdot)$ is the hashing operation. In HE based methods~\citep{muller2022instant, tancik2023nerfstudio}, especially in few view cases, the model may not able to learn how to correct the hash collision due to lack of training data. 
In typical rendering tasks, much of the space is empty, a phenomenon leveraged by HE-based methods for more efficient learning. However, without proper supervision, the model may fail to render unseen regions, as the HE transformation brings the input too close to known empty space. 
\begin{theorem}
In sparse scenario, where most of the space is empty,
\begin{equation}\label{eq:empty}
p(F(x;\theta)>0)\ll p(F(x;\theta)=0),~x\in\mathcal{X}
\end{equation}
where the RGB (final rendering) epistemic uncertainty caused by hash collision in HE cannot be detected by any ensemble based method, including \method, $\mathcal{X}$ is set of potential inputs. The ensemble based method refers to epistemic uncertainty estimation by the variance of ensembles $F(x;\theta+\delta)$, where
$|F(x,\theta) - \mathbb{E}_x[F(x;\theta+\delta)]|<\epsilon$, $\delta$ is the random perturbation.
\end{theorem}
\begin{proof}
We extend the ensembles to hash encoding case as 
\[|F(h(x);\theta) - \mathbb{E}_x[F(h(x);\theta+\delta)]|<\epsilon\]
Considering a significant portion of rendering output is empty, when hash collision happens on new input $x'$, we are very likely to have $F(h(x'),\theta)=0$ according to Eq.~\ref{eq:empty}. Because this is a rendering problem, the expected output must be positive, and so the ensembles $|\mathbb{E}_x[F(h(x');\theta+\delta)]|<\epsilon$, which means the output of ensemble is almost zero. Therefore, as long as hash collision happens and the scenario is sparse, the output would be likely to be zero without variance, i.e., 
\[
\mathrm{Var}[F(h(x');\theta+\delta)]=\frac{1}{N}\sum(F(h(x');\theta+\delta)-\Bar{F}(h(x');\theta+\delta))^2<\epsilon^2
\]
Near zero variance means no uncertainty, hence the epistemic uncertainty will not be reflected.
\end{proof}

Similarly, we can extend this conclusion to random initialization scheme. As long as the $h(\cdot)$ is a consistent pseudo hash function, the collision will happen at the same input, and will not give any informative information on the input with collision.

\textit{Takeaway}. 
Due to hash collisions, faithful epistemic uncertainty estimation (on RGB) is not feasible when hash collision happens. Epistemic uncertainty in depth prediction is influenced by hash collision as well. See extended discussion about depth prediction and overall uncertainty in \S\ref{ap:collision}.

\section{Performance Evaluation}\label{sec:eval}


\subsection{Evaluation on Tasks}
In rendering tasks, ground truth for epistemic uncertainty is unattainable, so we validate the effectiveness of uncertainty estimation indirectly through diverse tasks.
\begin{tightlist}
    \item \textbf{Active Learning}~\citep{gal2017deep, raj2022convergence, nguyen2022measure}: 
Correlation with Training Data Sufficiency $\rho_{\text{U}}$ and $\rho_{\text{R}}$. A faithful estimation of epistemic uncertainty should show a lower uncertainty with a more training views. Therefore we take $\rho_{\text{U}} = \rho_{\text{s}}(\overline{\sigma_{\text{max}}},N_{\text{train}})$ as metric, where $\rho_{\text{s}}(\cdot)$ represents the Spearman's Correlation~\citep{corder2014nonparametric}, $N_{\text{train}}$ is the number of training views, $\overline{\sigma_{\text{max}}}$ is defined in Alg.~\ref{alg:phd}. \textbf{$\rho_{\text{U}}$ is expected to be negative}. Similarly, the redundancy (dropout ratio $r_{\text{drop}}$ in Alg.~\ref{alg:phd}) should increase as more training view is available, i.e., better overfitting according to Theorem~\ref{theo:must_redund}. We introduce $\rho_\text{R} = \rho_s(r_{\text{drop}}, N_\text{train})$ as the other metric reflecting correlation between model robustness and training views. Hence \textbf{$\rho_\text{R}$ should be positive}. We streamline the estimation of the correlation coefficient by focusing on 8-view, 16-view, and 100-view setups, leveraging the overall trend for reasonable approximation.
    \item \textbf{Correlation with Prediction Error}~\citep{liu2019accurate, nannapaneni2016reliability}: $r_{\text{PE}}$. Since the rendering function in view synthesis is deterministic, the primary sources of prediction error are epistemic uncertainty and model mis-specification. Therefore, a correlation between uncertainty estimation and actual error is expected. Specifically, we have $\rho_{\text{PE}} = \rho_{\text{s}}(\zeta(x),\mathrm{RMSE}(F(x),F_{\text{GT}}(x)))$, where $\mathrm{RMSE}(\cdot)$ is the Root Mean Squared Error per pixel and channel, $F_{\text{GT}}(x)$ denotes the ground truth image.
\end{tightlist}



Besides  effectiveness of \method, we also highlight the efficiency of \method in Figure~\ref{fig:inference_speed}. 
Even when focusing solely on inference speed, without accounting for other practical constraints in alternative methods, \method demonstrates a performance gain of at least two orders of magnitude. This substantial efficiency improvement makes \method the only viable option for use during runtime, with only a minimal frame rate drop on the initial render.

\textbf{Datasets}: We conducted experiments to evaluate the performance of \method across 3 widely-used datasets: NeRF Synthetic Blender~\citep{mildenhall2021nerf}, Tanks \& Temples (T\&T)~\citep{Knapitsch2017} and the LLFF dataset~\citep{mildenhall2019llff}. 


\subsection{Active Learning}\label{subsec:al}
\begin{figure}
\centering
\includegraphics[width=\textwidth]{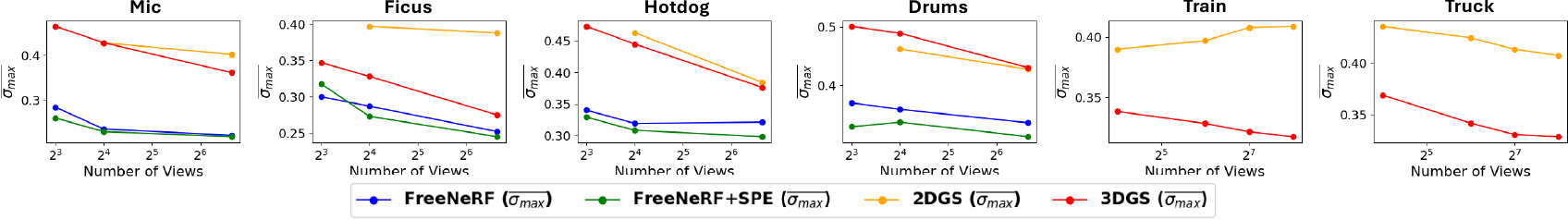}
\vspace{-0.8em}
  \caption{\textbf{Active Learning - $\overline{\sigma_{\text{max}}}$}: PH-Dropout robustness to active learning is showed by a  decreasing epistemic uncertainty at decreasing $\overline{\sigma_{\text{max}}}$, with increasing number of training views.}
  \label{fig:active_bounded_sigma}
\end{figure}

\begin{figure}
\centering
\includegraphics[width=\textwidth]{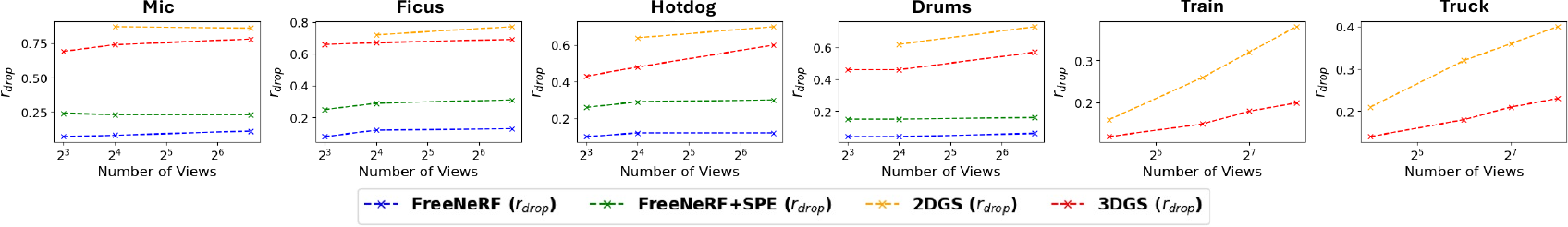}
\vspace{-0.8em}
  \caption{\textbf{Active Learning - $r_{\text{drop}}$}: PH-Dropout robustness to active learning is showed by a decreasing epistemic uncertainty at increasing $r_{\text{drop}}$, with increasing number of training views.}
  \label{fig:active_bounded_r_drop}
\end{figure}

For bounded case we evaluate the model on the Blender dataset following \cite{mildenhall2021nerf}. 
The baselines for the bounded case of NeRF include FreeNeRF~\citep{DBLP:conf/cvpr/YangPW23} and its fine-tuned variant FreeNeRF+SPE~\citep{DBLP:conf/icml/SunYXMSCM24}. These methods achieve superior fidelity in limited training view scenarios (few-view) and are free from hash collision issues, making them ideal benchmarks to demonstrate the effectiveness of \method. GS based method includes 3DGS~\citep{kerbl20233d} and 2DGS~\citep{DBLP:conf/siggraph/HuangYC0G24}, yielding better fidelity and efficiency than NeRF based methods when there are sufficient training views. The results are demonstrated in Figure~\ref{fig:active_bounded_sigma} and Figure~\ref{fig:active_bounded_r_drop} (detail in \S\ref{ap:al}, Table~\ref{table:active_bounded}), where the dropout ratio $r_{\text{drop}}$ increases as the number of training view increases, indicating higher redundancy of the trained model. Meanwhile, the models tend to have higher uncertainty $\overline{\sigma_{\text{max}}}$ when the number of training view decreases, even with smaller $r_{\text{drop}}$. 2DGS is the only outlier on the trend of $\overline{\sigma_{\text{max}}}$ because model collision in few view cases, \S\ref{ap:2dgs}.

FreeNeRF + SPE exhibits greater redundancy and lower UQ compared to FreeNeRF, despite both being based on NeRF with the same number of parameters. SPE~\citep{DBLP:conf/icml/SunYXMSCM24} simplifies function learning by altering just one activation function (discussed in detail in the \S\ref{ap:pespe}). This subtle structural change is clearly detected by \method, further demonstrating the effectiveness of \method.

For unbounded cases, we primarily focus on GS-based methods (2DGS and 3DGS), as conventional NeRF methods are too slow in this context without offering fidelity improvements. While HE-based NeRF is faster, it suffers from hash collisions in few-view setups, making it incompatible with other methods. Therefore, we explore NeRF for unbounded scenarios in a later section, where we implement \method only on unbounded NeRF with sufficient training views.
The results of GS-based models in unbounded scenarios, as shown in Figure~\ref{fig:active_bounded_sigma} and \ref{fig:active_bounded_r_drop} (detail in \S\ref{ap:al}, Table~\ref{table:GS_al_unbounded}), reveal a consistent pattern with the bounded scenarios. Specifically, as the number of training views increases, the dropout ratio $r_{\text{drop}}$ rises, while UQ metric $\overline{\sigma_{\text{max}}}$ decreases.

Combining results from both bounded and unbounded scenarios across NeRF and GS-based methods, we find a clear negative trend in $\rho_{\text{U}}$ and a significant positive trend in $\rho_{\text{R}}$. This demonstrates that the UQ provided by \method is well-suited for supporting active learning tasks. This effective UQ is then applied to an uncertainty-driven ensemble usecase in \S\ref{sec:usecase}.

\subsection{Correlation between uncertainty and prediction error}

\begin{figure}
\centering
  \includegraphics[width=0.95\textwidth]{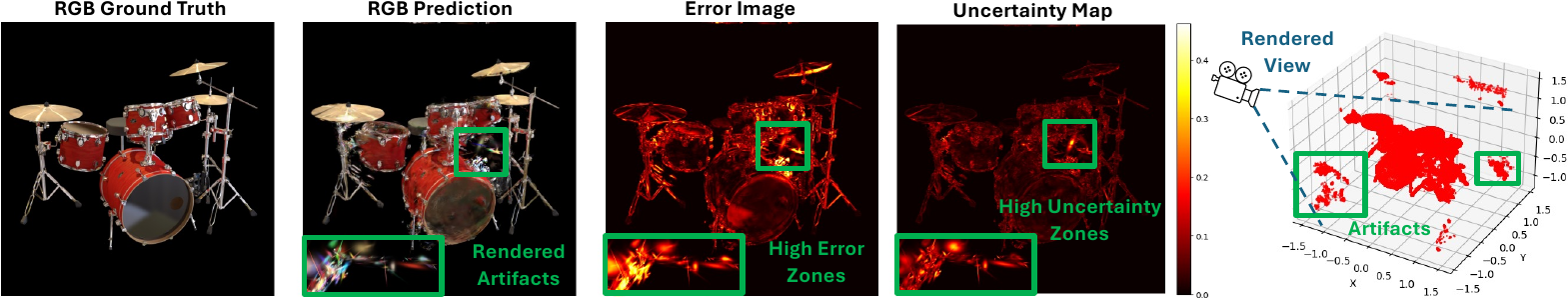}
  \vspace{-0.5em}
  \caption{Correlation between \method based epistemic uncertainty estimation and the actual RMSE, with 3DGS at bounded scenario (Blender drum). 8 training views.}
  \label{fig:drum_3dgs_8v}
  \vspace{-1em}
\end{figure}

\begin{figure}
\centering
\includegraphics[width=0.9\textwidth]{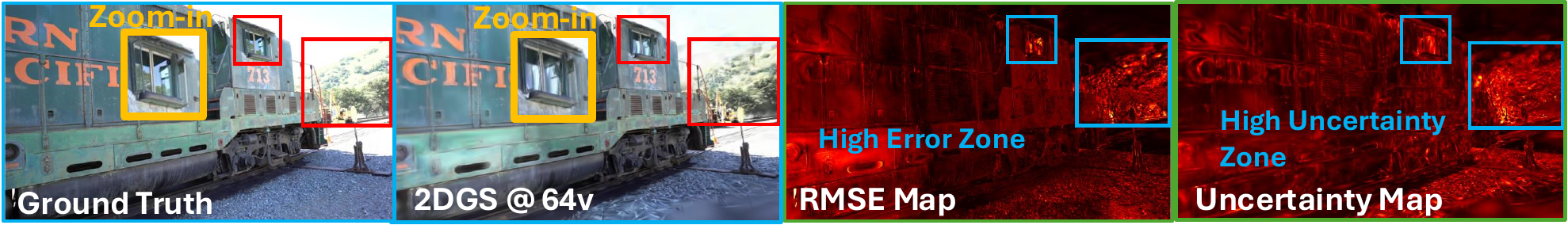}
\vspace{-0.5em}
  \caption{Correlation between \method based epistemic uncertainty estimation and the actual RMSE, with 2DGS at unbounded scenario. 64 training views.}
  \label{fig:2dgs_64v}
\end{figure}

\begin{figure}
\centering
\includegraphics[width=\textwidth]{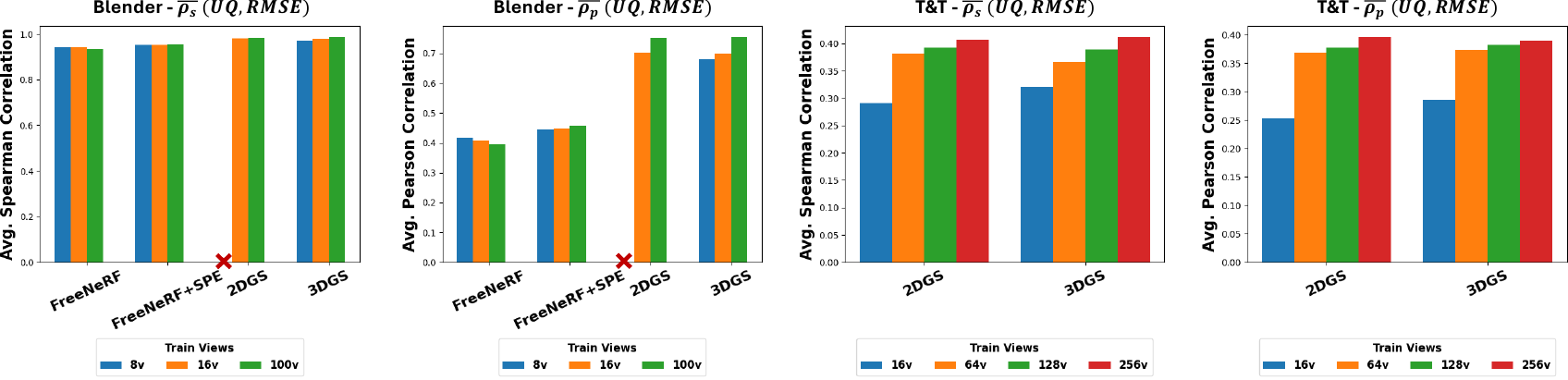}
\vspace{-1em}
  \caption{Correlation between RMSE and UQ of \method in bounded (Blender) and unbounded (T\&T) scenarios.}
  \label{fig:correlation_rmse_uq}
\end{figure}

As Figure~\ref{fig:drum_3dgs_8v} and Figure~\ref{fig:2dgs_64v} illustrate, in 2DGS and 3DGS, high RMSE region tends to overlap with high uncertainty region. The correlation between RMSE and UQ in different scenarios is demonstrated in Figure~\ref{fig:correlation_rmse_uq}. 
In bounded scenarios, both NeRF and GS-based methods demonstrate a strong correlation between RMSE and UQ. Results for 2DGS with 8 views are missing due to training limitations. In unbounded scenarios, GS-based models show lower correlation compared to bounded cases, reflecting the complexity of real-world data versus synthetic objects~\citep{ren2024nerf}. Additionally, both 2DGS and 3DGS exhibit higher correlation as training views increase, suggesting that insufficient training data leads to more unpredictable and random RMSE values. More detail about the correlation between \method UQ and RMSE is showed in \S\ref{ap:rho} Table~\ref{table:fid_bounded} for bounded cases, and Table~\ref{table:train_truck} for unbounded cases. 

In addition to the methods discussed in Figure~\ref{fig:correlation_rmse_uq}, we also explore hash encoding (HE) methods for error prediction, as they are not limited to few-view scenarios. For the unbounded case, we compare \method against Bayes Rays~\citep{goli2024bayes} using its NeRFacto setup~\citep{tancik2023nerfstudio}, which, to our knowledge, represents the current state-of-the-art for unbounded NeRF and serves as an enhanced version of InstantNGP~\cite{DBLP:journals/tog/MullerESK22}.

\begin{figure*}[ht]
\centering
\begin{minipage}[b]{0.65\textwidth}
    \centering
    \setlength{\tabcolsep}{4pt}
    \begin{small}
    \resizebox{\textwidth}{!}{%
        \begin{tabular}{l|c|c|c|c|c|c|c|c|c}
        \toprule
        Dataset & Method & PSNR $\uparrow$ & SSIM $\uparrow$ & $\rho_{\text{s}}$ $\uparrow$ & $\rho_{\text{p}}$ $\uparrow$ & AUSE RMSE $\downarrow$ & AUSE MSE $\downarrow$ & AUSE MAE $\downarrow$ \\
        \hline
        \multirow{2}{*}{Africa} 
        & Bayes Rays & 22.1 & 0.839 & 0.020 & -0.071 & 0.545 & 0.512 & 0.512 \\
        & PH-Dropout & 22.1 & 0.839 & \textbf{0.163} & \textbf{0.154} & \textbf{0.489} & \textbf{0.485} & \textbf{0.441} \\
        \hline
        \multirow{2}{*}{Basket} 
        & Bayes Rays & 22.8 & 0.823 & -0.335 & -0.241 & \textbf{0.410} & \textbf{0.304} & \textbf{0.287} \\
        & PH-Dropout & \textbf{22.9} & \textbf{0.825} & \textbf{0.342} & \textbf{0.310} & 0.438 & 0.345 & 0.351 \\
        \hline
        \multirow{2}{*}{Torch} 
        & Bayes Rays & 24.4 & 0.867 & -0.395 & -0.196 & 0.454 & \textbf{0.314} & 0.348 \\
        & PH-Dropout & \textbf{24.5} & 0.867 & \textbf{0.472} & \textbf{0.314} & \textbf{0.428} & 0.367 & \textbf{0.277} \\
        \hline
        \multirow{2}{*}{Statue} 
        & Bayes Rays & 19.9 & 0.813 & -0.469 & -0.285 & \textbf{0.369} & \textbf{0.187} & \textbf{0.216} \\
        & PH-Dropout & \textbf{20.0} & 0.813 & \textbf{0.370} & \textbf{0.166} & 0.596 & 0.469 & 0.468 \\
        \hline
        \multirow{2}{*}{Avg.} 
        & Bayes Rays & 22.30 & 0.836 & -0.295 & -0.198 & \textbf{0.445} & \textbf{0.329} & \textbf{0.341} \\
        & PH-Dropout & \textbf{22.38} & 0.836 & \textbf{0.337} & \textbf{0.236} & 0.488 & 0.417 & 0.384 \\
        \bottomrule
        \end{tabular}%
    }
    \end{small}
    \captionof{table}{\textbf{Comparison of Bayes Rays and \method on NeRFacto}: Bayes Rays fails to correlate depth uncertainty with high prediction error on the LF dataset.}
    \label{table:bayesrays_phdropout}
\end{minipage}
\hfill
\begin{minipage}[b]{0.33\textwidth}
    \centering
    \includegraphics[width=\textwidth]{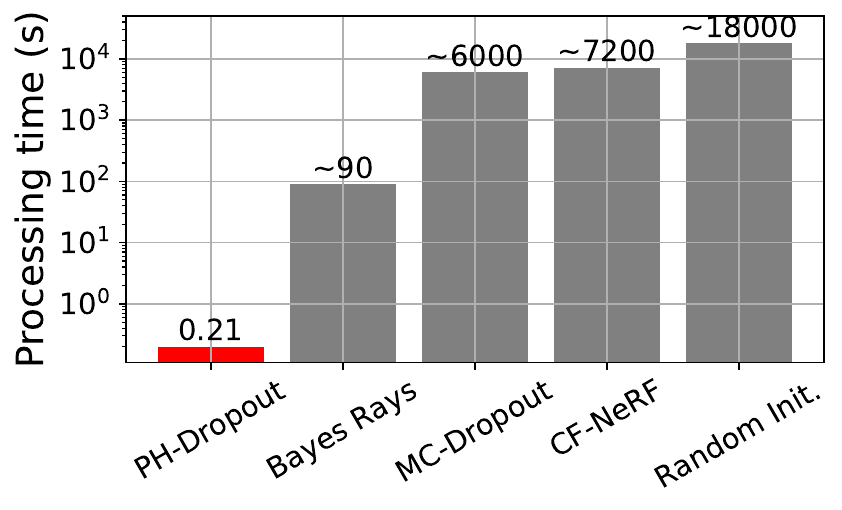}
    \vspace{-1em}
    \caption{\method can be applied on-the-fly to a trained method, yielding orders of magnitude efficiency gain.}
    \label{fig:inference_speed}
\end{minipage}
\end{figure*}

As for the comparison with baseline, we mainly focus on Bayes Rays~\citep{goli2024bayes}, which carries out depth uncertainty of NeRF. Related work as NeRF OTG~\citep{ren2024nerf} is not considered as a baseline as it is proposed for dynamic scenarios. CG-SLAM~\citep{hu2024cg} has recently explored similar spatial uncertainty-aware methods for GS models. However, it cannot be included either since CG-SLAM is not open-source and lacks sufficient implementation details. Computation-heavy methods like CF-NeRF~\citep{shen2022conditional} are excluded as they cannot be applied to more representative models, limiting the usefulness of their UQ. In Table~\ref{table:bayesrays_phdropout}, \method achieves higher correlation ($\rho_s$ and $\rho_p$) between UQ and RMSE. This comes at no fidelity cost, empirically validating the negligible impact of inference-only dropout due to expected high model redundancy (see Theorem \ref{theo:must_redund}).

\subsection{Usecases: Uncertainty Driven Model Ensembles}\label{sec:usecase}
Here we consider uncertainty driven model ensembling~\citep{wang2023diversity} as the usecase to further demonstrate the effectiveness of \method. The ensemble method is driven by selecting the image with lower overall uncertainty, \emph{i.e.}, select function $F$ from $F_a$ and $F_b$, following $\arg\min_{F\in\{F_a, F_b\}}\overline{\zeta_F(x))}$, where $\zeta_F(\cdot)$ is the per pixel and channel UQ under function $F$, $\overline{\zeta_F(x))}$ is the mean over all pixels and channel.

We randomly selected two non-overlapping 16-view training sets to train models `16v-a' ($F_a$) and `16v-b' ($F_b$). This task is challenging due to the random selection and the proximity between views in both sets, requiring the model to be highly sensitive in choosing the correct rendering results. As an ensemble method, we expect \method to select the optimal view, ensuring overall fidelity that matches or exceeds the best ground-truth model between the two models.

Metric $E_\text{ME}$ is introduced to quantify the performance of ensembling. Here we consider model ensembling with dynamic selection, where two models with exactly same configuration are trained with different views of the same object, denoting as $F_a$ and $F_b$. We aim to evaluate the expected value of the following ratio, which directly reflects \method's effectiveness in estimating uncertainty due to insufficient training views
\[E_\text{ME} = \mathbb{E}(r_\text{ME})=\mathbb{E}_x\left(\frac{\text{SSIM}(F(x)|\arg\min_{F\in\{F_a, F_b\}}\overline{\zeta_F(x))}}{\max(\text{SSIM}(F_a(x),F_{\text{GT}}(x)),\text{SSIM}(F_b(x),F_{\text{GT}}(x)))}\right)\]
Intuitively, if the estimation can guide the selection of more suitable model, we will have $E_{\text{ME}}\rightarrow1$.

The detailed results are included in \S\ref{ap:usecase}, Table~\ref{table:uncertainty_drive_ensemble} for bounded cases, where \method successfully selects most correct views in 2DGS, with the ensemble model consistently performing at or near the level of the model with better training views. In 3DGS, \method consistently selects the correct view, with $E_{\text{ME}}$ always close to 1. \method performs better in 3DGS primarily because 2DGS experiences collision issues in few-view scenarios, leading to missing renderings, similar to hash collisions in HE-based methods. (see \S\ref{ap:2dgs}).

\S\ref{ap:usecase} Table~\ref{table:uncertainty_drive_ensemble_unbound} shows the ensemble performance in unbounded cases. \method effectively selects images with superior fidelity, enabling the ensemble model to outperform any individual model. Similar to the bounded case, \method shows reduced performance on 2DGS due to the inherent limitations of the 2DGS method.

\begin{table*}
\centering
\setlength{\tabcolsep}{4pt}
\begin{small}
\resizebox{0.65\columnwidth}{!}{%
    \begin{tabular}{l|c|c|c|c|c|c|c|c|c|c}
    \toprule
    Method & \multicolumn{5}{c}{2DGS} & \multicolumn{5}{c}{3DGS} \\
    \hline
    Dataset 
    &Metric & 16v-a & 16v-b  &selected & $E_\mathrm{ME}$ & 16v-a & 16v-b &selected & $E_\mathrm{ME}$\\
    \hline
    \multirow{2}{*}{Blender} 
    & SSIM & \textbf{0.887} & 0.873 & \underline{0.881} & 0.974 & \underline{0.900} & 0.894 & \textbf{0.912} & 0.996\\
    & PSNR & \textbf{24.4} & 24.1 & \underline{24.3} & & \underline{25.4} & 25.3 & \textbf{26.3} & \\
    \hline
    \multirow{2}{*}{T\&T} 
    & SSIM & \underline{0.564} & 0.553 & \textbf{0.591} & 0.952 & \underline{0.550} & 0.537 & \textbf{0.601} & 0.987\\
    & PSNR & 15.4 & \underline{15.6} & \textbf{16.3} & & 15.8 & 15.8 & \textbf{17.1} & \\
    \bottomrule
    \end{tabular}%
}
\end{small}
\vspace{-0.7em}
\caption{Performance on selecting rendered views with the lowest $\overline{\sigma_{\text{max}}}$ from an ensemble of two models with different training views.}
\label{table:uncertainty_drive_ensemble_both}
\end{table*}



\section{Discussion}

\subsection{Different Performance on NeRF and GS, Different Encoding Methods}
\begin{figure}
\centering
\includegraphics[width=0.8\textwidth]{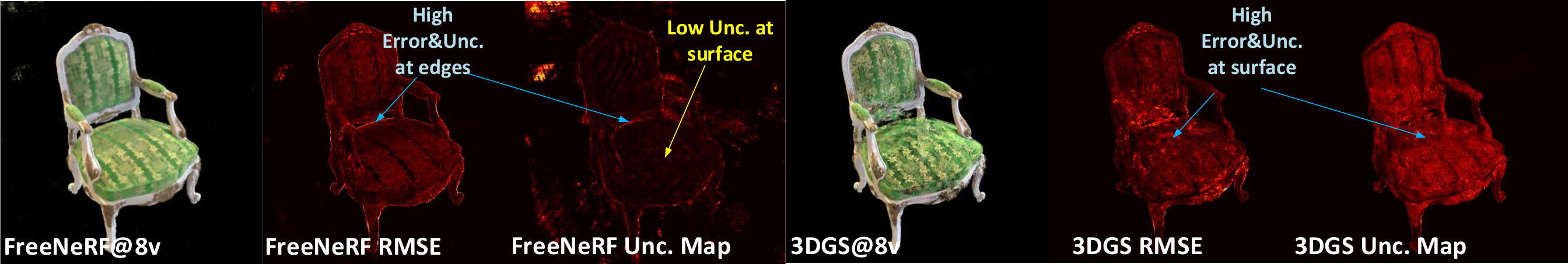}
  \caption{NeRF and GS models show distinct features in UQ and RMSE.}
  \vspace{-0.5em}
  \label{fig:gs_vs_nerf}
\end{figure}

Throughout this paper, NeRF and GS-based methods exhibit distinct patterns in redundancy and correlation with RMSE. As indicated by Theorem~\ref{theo:must_redund}, these differences arise from the fundamental ways each method approximates the rendering function. Figure~\ref{fig:gs_vs_nerf} illustrates that NeRF tends to show higher error and uncertainty at object edges, while GS models display increased uncertainty on smooth surfaces. This behavior aligns with the theorem and is a key factor behind the distinct UQ performance of each method.

\subsection{Limitations of \method}

\method struggles with UQ in the presence of input hash collisions or similar model collisions in 2DGS, limiting its applicability to hash encoding-based methods~\citep{muller2022instant, tancik2023nerfstudio}. Additionally, \method is specifically designed for view synthesis tasks; further research is required to adapt it for other applications.

\section{Conclusion}

We present \method, an efficient and effective epistemic uncertainty quantification (UQ) method for view synthesis, designed to operate directly on trained models. \method is compatible with both NeRF and GS-based methods and can be applied to both bounded objects and unbounded scenarios. By offering fast inference and easy implementation, \method makes epistemic UQ practical and stands as the first training-free method for UQ in GS models. Extensive evaluations across a broad range of downstream applications highlight its effectiveness. Theoretical analysis of \method also uncovers fundamental differences and connections between NeRF and GS rendering methods, paving the way for future research to enhance their efficiency, fidelity, and scalability.

\bibliography{iclr2024_conference}
\bibliographystyle{iclr2024_conference}

\appendix
\section{Appendix}
\subsection{Compute the std of a multi-channel image}\label{ap:sigma}
Here we explain how to define the overall variance of the image $x$.

Let $\zeta(x) = \sigma \in \mathbb{R}^{N \times H \times W \times C}$ be a tensor representing the standard deviation image $x$ of $N$ views after $S$ stochastic forward-passes using \method. The dimension of the rendered sampled images is $H \times W \times C$, where $H$ is the height, $W$ is the width, and $C$ is the number of channels.

For each view $i \in \{1, 2, \dots, N\}$, we define the maximum standard deviation as:
\[
\sigma_{\text{max}, i} = \max_{h, w, c} \left( \sigma_{i, h, w, c} \right)
\]
where $h \in \{1, 2, \dots, H\}$, $w \in \{1, 2, \dots, W\}$, and $c \in \{1, 2, \dots, C\}$.

We use the maximum value instead of the mean due to the sparse nature of the uncertainty map and the overall rendering process. A large portion of the pixels or space in the rendering is either empty or easily predictable, making the mean value ineffective for capturing meaningful variations, thus reducing sensitivity in quantification.

The mean of the maximum standard deviations across all $N$ views, denoted as $\overline{\sigma_{\text{max}}}$, is defined as:
\[
\overline{\sigma_{\text{max}}} = \frac{1}{N} \sum_{i=1}^{N} \sigma_{\text{max}, i} = \frac{1}{N} \sum_{i=1}^{N} \max_{h, w, c} \left( \sigma_{i, h, w, c} \right)
\]

\subsection{MC-dropout is not suitable for NeRF}\label{ap:mc}
Overall, training dropout is a linear approximation of the average of ensembles, and hence prevent overfitting~\citep{srivastava2014dropout}. The target rendering function is unique and deterministic, hence this technology cannot bring performance gain but only reduce the training efficiency and approximation accuracy.
\begin{theorem}\label{theo:notMCD}
Training dropout prevents the efficient convergence of NeRF MLP on the rendering function. 
\end{theorem}

\begin{proof}
The color to render a ray $r$, \emph{i.e.}, the rendering function is defined as:
\[
C(r) = \int_{t_n}^{t_f}T(r)\sigma(r(t))c(r(t),d)dt
\]
where $T(r) = \exp{-\int^t_{t_n}\sigma(r(s))ds}$ denotes the accumulated transmittance along the ray from from $t_n$ to $t$. Considering the typical NeRF with PE approximate rendering function using Fourier features should have a unique spectrum as: $C(r) = \sum w_i(r)\sin(r)$, $w_i(r)$ is the Fourier features. Applying dropout is equal to remove a few features, \emph{i.e.}, $C'(r) = \rho\sum\mu_i w_i(r)\sin(r)$, where $\mu_i\sim \mathrm{Bern}(p, N_f)$, $\rho = 1/(1-p)$. Since they both converge on training set, then on training set:
\begin{equation} 
|\sum w_i(r)\sin(r) - \rho\sum\mu_i w_i(r)\sin(r)|<\epsilon
\end{equation}
when $p=0$ (no dropout), we have $\mu_i=1$ as solution. However, when $p\neq 0$, an approximation can be made under certain conditions (ignore the empty space): (1) $p$ is small, so the dropout rate is low, ensuring that the majority of neurons remain active and the model behavior closely approximates the no-dropout scenario; (2) the power of each component is distributed sparsely, meaning the dropped components contribute minimally to the overall output, or the components exhibit even power distribution, akin to white noise.

Since the rendering function in most practical scenarios is not equivalent to white noise, it follows that the dropout ratio must remain small by default to avoid excessive loss of important information. Additionally, the distribution of power across components tends to be sparse in real-world cases, implying that only a few components carry significant influence.

This suggests that models with dropout can only effectively approximate cases where the components are sparse, leading to patterns that are simpler and lack fine-grained detail. As a result, while dropout helps prevent overfitting, it may also limit the model's capacity to capture intricate patterns when too much information is dropped.
\end{proof}


Empirically, previous works \cite{shen2024estimating, sunderhauf2023density} have proven that the estimation of MC-dropout on NeRF is inaccurate with significant worse rendering quality when trained with dropout.

\subsection{Condition of Effective uncertainty estimation}\label{ap:effective}
\begin{theorem}\label{thm:ensemble}
Suppose instances $F$ in a set of models $\hat{\mathcal{F}}$ (trained on same data $\mathcal{D}$), \emph{i.e.},
\[
\forall \hat{\mathcal{F}} \subset \mathcal{F}, |\hat{\mathcal{F}}| > T \rightarrow \frac{\sum_{\forall F \in \mathcal{F}}\hat{y}_F}{{|\mathcal{F}|}} = \frac{\sum_{\forall F \in \hat{\mathcal{F}}}\hat{y}_F}{|\hat{\mathcal{F}}|} + \epsilon_{\text{mean}}
\]

exhibit identical and perfect fitting performance on the training set, where $\hat{y}_F = F(x)$, $\epsilon_{\text{mean}}$ is the approximation error. The variation of their output can be interpreted as a reflection of epistemic uncertainty, if under a random initialization scheme $S(\cdot)$, 
\[
\frac{|\hat{\mathcal{F}}\cap S(\mathcal{F})|}{|S(\mathcal{F})|} = \frac{|\hat{\mathcal{F}}\cap \mathcal{F}^*|}{|\mathcal{F}^*|}\gg 0
\]
where $\mathcal{F}^* = S(\mathcal{F})$ is the result of random initialization, assuming the data is non-noisy and deterministic.
\end{theorem}

\begin{proof}
Let $\mathcal{F}$ be the set of all possible models that could explain the dataset $\mathcal{D}$. Each model $F\in\mathcal{F}$ provides a prediction $\hat{y}$ for given input $x$. We define function $F$ can explain a dataset if and only if
\[
\forall (x, y) \in \mathcal{D} \rightarrow |y - F(x)| = |y - \hat{y}| \leq \epsilon
\]
where $y$ is the expected and deterministic output to input $x$, appearing in pairwise in dataset $\mathcal{D}$

The predictive distribution of the model can be expressed as:

\[
p(\hat{y}|x,\mathcal{D}) = \int_{\mathcal{F}}p(\hat{y}|x,F)p(F|\mathcal{D})dF
\]

The epistemic uncertainty can be then represented as 
\[
\mathrm{Var}[\hat{y}|x,\mathcal{D}]=\int_\mathcal{F} p(F|\mathcal{D})(\hat{y}_F-\mathbb{E}[\hat{y}|x,\mathcal{D}])^2 dF
\]

where $\hat{y}_{F} \sim p(\hat{y}|x,F)$ and $\mathbb{E}[\hat{y}|x,\mathcal{D}]$ is the expectation of the prediction over the model posterior distribution.

To estimate the value of uncertainty $\mathrm{Var}[\cdot]$, we can conduct a Monte-Carlo solution. We first obtain an unbiased and significant number of instance $F$ forms set $\Tilde{\mathcal{F}}$, and $\Tilde{\mathcal{F}}\subset\mathcal{F}$. The expectation of prediction is estimated as $\frac{\sum_{\forall F \in \Tilde{\mathcal{F}}}\hat{y}_F}{|\Tilde{\mathcal{F}}|}$. The the estimated uncertainty is 
\[
\Tilde{\mathrm{Var}}[\hat{y}|x,\mathcal{D}] = \sum_{F \in \Tilde{\mathcal{F}}} \frac{N_F}{|\Tilde{\mathcal{F}}|}(\hat{y}_F-\frac{\sum_{\forall F \in \Tilde{\mathcal{F}}}\hat{y}_F}{|\Tilde{\mathcal{F}}|})^2 = \mathrm{Var}[\hat{y}|x,\mathcal{D}] + \epsilon_{\text{APP}}
\]

\[
\lim_{|\Tilde{\mathcal{F}}|\rightarrow+\infty}\Tilde{\mathrm{Var}}[\hat{y}|x,\mathcal{D}] = \mathrm{Var}[\hat{y}|x,\mathcal{D}]
\]

where $\epsilon_{\text{APP}}$ is the error caused by the bias of limited sampling number.

So far we have present the ideal case of uncertainty estimation. To obtain an ideal $\Tilde{\mathcal{F}}$ is difficult because of the computation overhead and bias in sampling (\emph{e.g.}, only consider $F$ with certain number of parameters). Here we discuss the feature of a subset $\hat{\mathcal{F}}\subset\Tilde{\mathcal{F}}$. 

Suppose the prediction expectation with $\hat{\mathcal{F}}$ has an error $\beta$ 

\[
\frac{\sum_{\forall F \in \Tilde{\mathcal{F}}}\hat{y}_F}{|\Tilde{\mathcal{F}}|} = \frac{\sum_{\forall F \in \hat{\mathcal{F}}}\hat{y}_F}{|\hat{\mathcal{F}}|} + \beta(x)
\]

the probability density of $F$ should be calibrate by $\alpha_F$

\[
p(F|\mathcal{D}) = \frac{\alpha_FN_F}{|\hat{\mathcal{F}}|}
\]

now the target estimation could be described with

\[
\Tilde{\mathrm{Var}}[\hat{y}|x,\mathcal{D}] = \sum_{F \in \hat{\mathcal{F}}} \frac{\alpha_FN_F}{|\hat{\mathcal{F}}|}(\hat{y}_F-\frac{\sum_{\forall F \in \hat{\mathcal{F}}}\hat{y}_F}{|\hat{\mathcal{F}}|}-\beta(x))^2
+ \sum_{F \in \Tilde{\mathcal{F}}-\hat{\mathcal{F}}} \frac{N_F}{|\Tilde{\mathcal{F}}|}(\hat{y}_F-\frac{\sum_{\forall F \in \Tilde{\mathcal{F}}}\hat{y}_F}{|\Tilde{\mathcal{F}}|})^2
\]

\[
 = \sum_{F \in \hat{\mathcal{F}}} \frac{\alpha_FN_F}{|\hat{\mathcal{F}}|}(\hat{y}_F-\frac{\sum_{\forall F \in \hat{\mathcal{F}}}\hat{y}_F}{|\hat{\mathcal{F}}|}-\beta(x))^2
+ \delta(x)
\]

\[
= \sum_{F \in \hat{\mathcal{F}}} \frac{\alpha_FN_F}{|\hat{\mathcal{F}}|}(\hat{y}_F-\gamma(x)-\beta(x))^2
+ \delta(x)
\]

\[
= \sum_{F \in \hat{\mathcal{F}}} \frac{\alpha_FN_F}{|\hat{\mathcal{F}}|}(\hat{y}_F-\gamma(x)-\epsilon)^2
+ \delta(x)
\]

where $\delta(\cdot)\geq0$, $\beta(\cdot)$ is determined by $\hat{\mathcal{F}}$ and $x$, $\gamma(x) = \frac{\sum_{\forall F \in \hat{\mathcal{F}}}\hat{y}_F}{|\hat{\mathcal{F}}|} = \frac{\sum_{\forall F \in \hat{\mathcal{F}}}F(x)}{|\hat{\mathcal{F}}|}\geq0$, and $0<\alpha_F$.

The estimation on $\hat{\mathcal{F}}$ without calibration is

\[
\hat{\mathrm{Var}}[\hat{y}|x,\mathcal{D}] = \sum_{F \in \hat{\mathcal{F}}} \frac{N_F}{|\hat{\mathcal{F}}|}(\hat{y}_F-\gamma(x))^2
\]
this is what we can compute directly.

Because all of the function in $\hat{\mathcal{F}}$ is equivalent to functions in $\mathcal{F}$, and $\Tilde{\mathcal{F}}$ is also a subset of $\mathcal{F}$, then we have $\beta(x)\rightarrow \epsilon$ when $|\hat{\mathcal{F}}|\rightarrow +\infty$.

We can compute $\alpha_F$ as $|\hat{\mathcal{F}}|/|\Tilde{\mathcal{F}}|$ if the size of both sets are available. The ratio of space $\alpha_F$ represents how the actual measurement contributes the ground truth uncertainty. And the actual uncertainty will be $\alpha_TV+\delta(x)$, as long as $\alpha_T \gg 0$, $V$ is an effective estimation, as large $V$ indicates high model uncertainty for sure.

Now we need to measure $|\hat{\mathcal{F}}|/|\Tilde{\mathcal{F}}|$. Given the complexity of the space of the high dimension functions, we cannot easily compute the exact value. However, we can still verify the $\alpha_T$ is not a negligible small value by doing sparse sampling over $\Tilde{\mathcal{F}}$ following reference schemes. Suppose there is a reference sampling scheme to obtain $\mathcal{F}^*$, if $1>\frac{|\hat{\mathcal{F}}\cap \mathcal{F}^*|}{|\mathcal{F}^*|}\gg 0$, then the uncertainty measurement on $\hat{\mathcal{F}}$ represents the a significant part of uncertainty on $\mathcal{F}^*$. Also because of $1>\frac{|\mathcal{F}^*|}{|\Tilde{\mathcal{F}}|}\gg0$, we have 

\[
\alpha_T = \frac{|\hat{\mathcal{F}}|}{|\Tilde{\mathcal{F}}|} > \frac{|\hat{\mathcal{F}}\cap \mathcal{F}^*|}{|\mathcal{F}^*|}\cdot \frac{|\mathcal{F}^*|}{|\Tilde{\mathcal{F}}|}\gg 0
\]

By default, we obtain $\mathcal{F}^*$ via random initialization, which has been proven to be an effective way to represent the model uncertainty. 

As a special case, if the data uncertainty at input $x$ is zero, then we must have
\[
\delta(x)=0, \forall F \in \hat{\mathcal{F}}, F(x) = \gamma(x)
\]

This means, if the uncertainty is very low, then given arbitrary $\hat{\mathcal{F}}$ with a significant size $\frac{|\hat{\mathcal{F}}|}{|\Tilde{\mathcal{F}}|}\gg0$, we should have a stable $F(x)$.

If the model shows high uncertainty at input $x$ within set $\hat{\mathcal{F}}$, then this uncertainty will contribute to a significant part of the ground truth uncertainty. And if the model has overall low uncertainty at $x$, $F(x)-\gamma(x) \approx 0, \forall F \in \hat{\mathcal{F}}$
\end{proof}

Random initialization is expensive and we cannot obtain $\mathcal{F}^*$ easily. Each trained model is just one instance of $\mathcal{F}^*$. Following aforementioned theorem, if we can find a subset of $\mathcal{F}^*$ with significant probability density, and guarantee the expectation of $\hat{y}$ converges to the global expectation with marginal error, then the estimation on this subset can reflect the lower bound of uncertainty.

\subsection{Explanation of difference between SPE and PE}\label{ap:pespe}
Here we prove that conventional positional encoding (PE)~\citep{tancik2020fourier} needs more parameters to approximate the same function than sinusoidal positional encoding (SPE)~\citep{DBLP:conf/icml/SunYXMSCM24}. We first investigate the following question: how to use signal of frequency $f_1$ and $f_2$ to create a new frequency component. Suppose signal $y(t)$ is the weighted combination of sinusoidal signal with frequency $f_1$ and $f_2$, representing the initial fully connected layer with PE, $A$ is the amplitude factor, and $F(\cdot)$ denotes the rest neural networks. To create a new frequency $\frac{f_1+f_2}{2}$, the following layers need performs a function $F^*(\cdot)$ to eliminate frequency $\frac{f_1-f_2}{2}$, as follows (because the signal is a combination of unique set of frequency features, $\frac{f_1-f_2}{2}$ may not be needed).

\begin{align}
F(y(t)) &= F(A\cos(2\pi\frac{f_1-f_2}{2}t)\sin(2\pi\frac{f_1+f_2}{2}t))\\  &=F^*(A\sin(2\pi\frac{f_1+f_2}{2}t))   
\end{align}

\begin{equation}
    F^*(\cdot) = \cos(2\pi\frac{f_1-f_2}{2}t)^{-1}F(\cdot)
\end{equation}

\begin{equation}
    f = \sum^{L}_{i=0}w_if_i, \text{where: }\sum w_i = 1, w_i \in \{m/2^n\}, m , n \in \mathbb{N}
\end{equation}

Obviously, the highest freuqncy can be represented is bounded by

\begin{equation}
    f \leq f_L = 2^{L-1}
\end{equation}

When $n$ is large enough, the error is bounded by $1/2^n$. Therefore, in theory, under ideal convergence, NeRF with PE can approximate arbitrary frequency within $2^{L-1}$ effectively.

However, when try to fine tune the high frequency features, the following effect will happen. The artifacts can be only be reduced when $f_1$ and $f_2$ are close. $f_{l} - f_{l-1} = f_{l-1}$ could be still require to learn high frequency representation directly via MLP. This results in NeRF is always struggling to approximate high frequency detail until the input sample rate is high enough (many views). 

If $\frac{f_1 + f_2}{2}$ is a new frequency features, then $\frac{|f_1 - f_2|}{2}$ is a new frequency feature as well. 

Following the standard PE, the input frequency component can be represented by $2^{L-1}$, the new created features is $(2^{n}+1)2^{m-1}$, so the synthetic frequency is always as $\text{Odd Num.}\times 2^n$ in pairwise.

    \begin{align}
       \frac{2^m+2^m\cdot2^n}{2}& = 2^{m-1}+2^{m-1}\cdot2^n \\
        &= (2^n+1)2^{m-1}
    \end{align}

in $\cos(\cdot)^{-1}$ side

    \begin{align}
       \frac{2^m\cdot2^n-2^m}{2} = (2^n-1)2^{m-1}
    \end{align}

suppose now merge with $k$

    \begin{align}
       \frac{(2^n+1)2^{m-1}+2^k}{2} = (2^n+1+2^{k-m+1})\cdot2^{m-2}
    \end{align}

    \begin{align}
       \frac{(2^n+1)2^{m-1}-2^k}{2} = (2^n+1-2^{k-m+1})\cdot2^{m-2}
    \end{align}

\subsection{Over-confident with \method in Hash Encoding}
This discussion tries to explain two phenomenon: (1) hash encoding based NeRF performs terrible in few-view cases, mixing object and background; (2) Hash encoding is more robust to \method, \emph{i.e.}, our method is less effective on hash encoding based NeRF.

In the hash encoding used in InstantNGP~\citep{muller2022instant} and NeRFacto~\citep{tancik2023nerfstudio}, the input is encoded in a pseudo-random manner. 
\begin{equation} \label{eq:hash-encoding}
   h(\mathbf{x}) =  \left(\bigoplus^d_{i=1}x_i\pi_i\right) \mod T
\end{equation}
where $\oplus$ denotes the bit-wise XOR operation and $\pi_i$ are unique,
large prime numbers, $T$ is the size of hash table. Due to the pseudo-random encoding, similar encoded values do not necessarily reflect spatial proximity. Without targeted supervision, the MLP tends to regress based on the absolute values of the encoding, often producing similar results for close encodings. This occurs because the MLP, with ReLU (or other continuous activations), exhibits smoothness and continuity~\citep{nair2010rectified,yarotsky2018optimal}, making it difficult to effectively distinguish between background and object in few-view InstantNGP and NeRFacto.

\textit{Takeaway}. In this paper, we apply \method to hash encoding \textbf{only in many-view and unbounded scenarios} due to its overconfidence issue and poor generalization to unseen views.

\subsection{Supplemental Results with Hash Encoding based Methods}
We present a comprehensive comparison of three methods: NeRF + HE, NeRF + PE, and GS, using the Blender dataset as a benchmark. We evaluate their performance across two key scenarios: (1) rendering fidelity in a few-view setup (8 views and 16 views), and (2) training speed with a sufficient number of training views. These scenarios address two critical aspects of view synthesis approaches: how well the model generalizes when observations are limited, and how efficient the training and inference processes are when ample training data is available.

We observe a significant performance improvement in the few-view setup with the PE + NeRF method. This gain is primarily due to the method's ability to learn a continuous function, allowing it to capture low-frequency, large-scale features that generalize effectively across the spatial domain. In the second task, all methods show high fidelity rendering performance given sufficient training view. GS achieves highest training and inference efficiency.

For HE based methods, we notice a significant drop when implement them on PyTorch only. This is because HE based method uses indexing operations  (\emph{e.g.,} look up after hashing), which is easier to be accelerated with c++ compiler. As a result, a significant portion of efficiency gain of HE is caused by the difference between PyTorch and c++. And even under the ideal setup with c++, it is still slower than GS model.

\textit{Takeaway}. HE-based methods receive less attention in this paper due to their poor performance in few-view setups and inefficiency in many-view scenarios.

\subsection{Supplementary Experiments: Unable to handle hash collision in hash encoding based methods}\label{ap:collision}
\begin{figure}
\centering
  \includegraphics[width=0.95\textwidth]{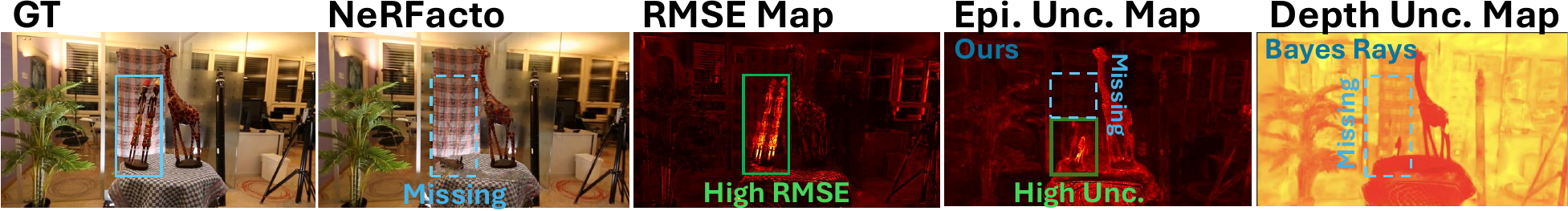}
  \caption{NeRFacto~\citep{tancik2023nerfstudio} fails to render the ``dolls'' due to hash collision. \method reveals part of the missing dolls but cannot render the fully collapsed part. Depth Uncertainty wit Bayes Rays~\citep{goli2024bayes}}
  \label{fig:nerfacto_bayes}
\end{figure}

\begin{figure}
\centering
  \includegraphics[width=0.9\textwidth]{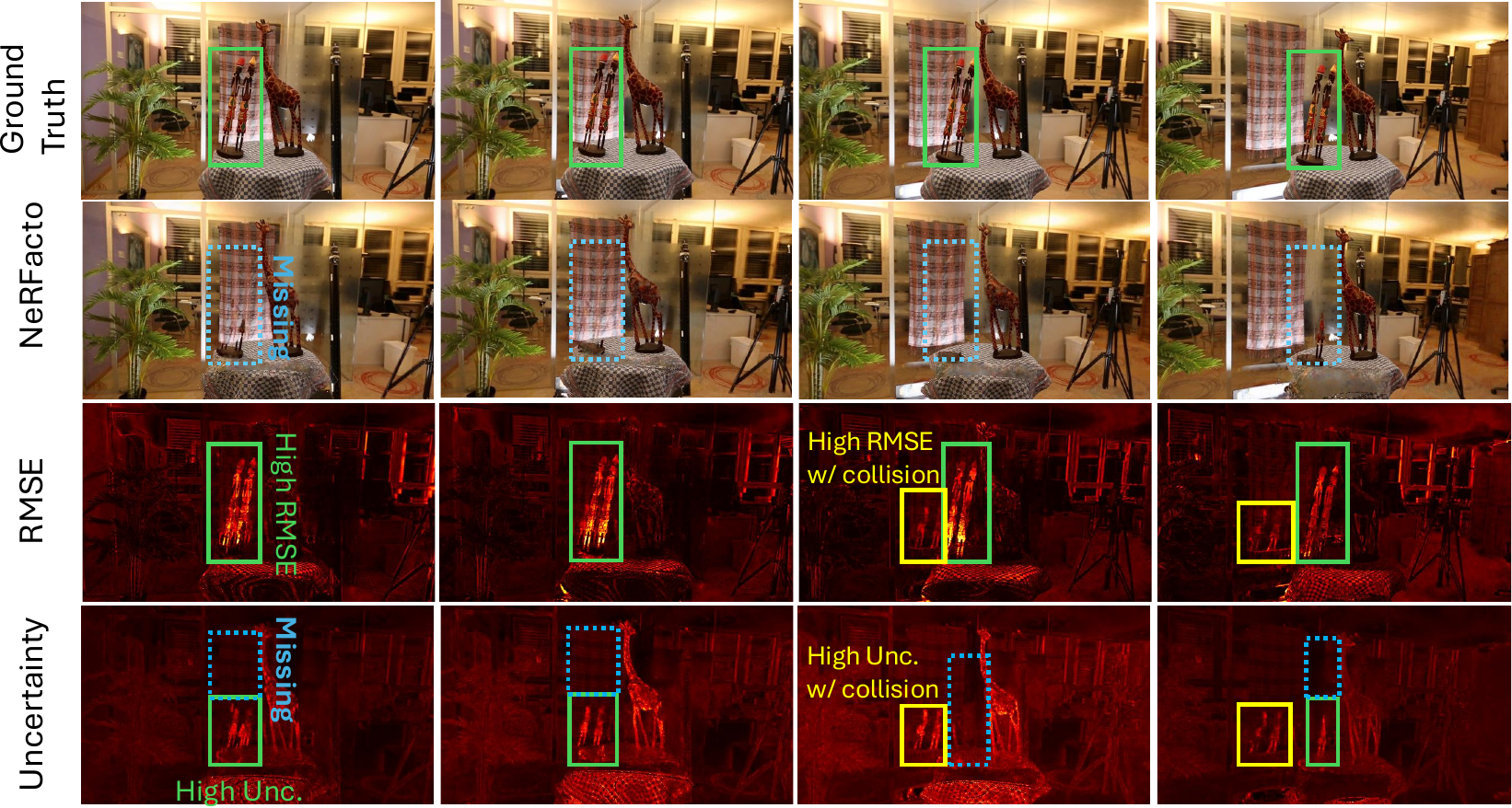}
  \caption{More views of \method performance under hash collision. \method can only track the rendered part, shift to left due the collision, highlighted with yellow}
  \label{fig:nerfacto_multiview}
\end{figure}

In Figure~\ref{fig:nerfacto_bayes} we show an example where NeRFacto does not render some objects, \emph{e.g.}, the dolls highlighted with blue. \method is able to show high uncertainty on the place NeRFacto tends to render but cannot show anything on the fully collapsed place. \method still yields better robustness when collision happens, because the other methods require training as Bayes Rays will experience collision issue more significantly and fail to render anything on the collapsed regions.
Figure~\ref{fig:nerfacto_multiview} further demonstrates the influence of hash collision in HE. The yellow ``ghost'' effect is replication of the dolls, the NeRFacto model mix up two different input, and it does not consistent on the spatial domain because the collision is pseudo random.

\subsection{Detailed results of active learning task}\label{ap:al}
Here we enclosed the detailed experiment results of the active learning usecase in \S\ref{subsec:al}, refer to the Table~\ref{table:active_bounded} and Table~\ref{table:GS_al_unbounded}.
\begin{table*}
\centering
\setlength{\tabcolsep}{4pt}
\begin{small}
\resizebox{0.75\columnwidth}{!}{%
    \begin{tabular}{l|c|c|c|c|c|c|c|c|c|c|c|c}
    \toprule
    Method & \multicolumn{4}{c}{FreeNeRF} & \multicolumn{3}{c}{FreeNeRF+SPE} & \multicolumn{2}{c}{2DGS} & \multicolumn{3}{c}{3DGS}\\
    \hline
    \multirow{2}{*}{Metric} & \multicolumn{12}{c}{Viewpoints (8v, 16v, 100v)} \\
    \cline{2-13} 
    & & 8v & 16v & 100v & 8v & 16v & 100v & 16v & 100v & 8v & 16v & 100v \\
    \hline
    \multirow{2}{*}{Mic} 
    & $\overline{\sigma_{\text{max}}}$ &0.283 &0.234 & 0.220 & 0.259 & 0.228 &0.217 & 0.428 & 0.402 & 0.465 & 0.428 & 0.361 \\
    & $r_{\text{drop}}$ &0.07 & 0.08& 0.11 & 0.24 & 0.23 &0.23 & 0.87 & 0.86 & 0.69 & 0.74 & 0.78 \\
    \hline
    \multirow{2}{*}{Chair} 
    & $\overline{\sigma_{\text{max}}}$ & 0.406 & 0.345 & 0.334 & 0.328 & \textcolor{red}{0.338} &0.315 & 0.470 & 0.424 & 0.487 & 0.451 & 0.377 \\
    & $r_{\text{drop}}$ & 0.08 & 0.07 & 0.10 & 0.19 & 0.19 &0.20 & 0.60 & 0.60 & 0.47 & 0.50 & 0.55 \\
    \hline
    \multirow{2}{*}{Ship} 
    & $\overline{\sigma_{\text{max}}}$ & 0.339 & 0.284 & \textcolor{red}{0.288} & 0.306 & \textcolor{red}{0.313} &0.298 & 0.427 & 0.417 & 0.430 & 0.382 & 0.346 \\
    & $r_{\text{drop}}$ & 0.09 & 0.10 & 0.08 & 0.23 & 0.23 &0.24 & 0.38 & 0.62 & 0.21 & 0.24 & 0.37 \\
    \hline
    \multirow{2}{*}{Materials} 
    & $\overline{\sigma_{\text{max}}}$ & 0.280 & 0.230 & 0.223 & 0.261 & 0.230 &0.211 & 0.426 & \textcolor{red}{0.436} & 0.502 & 0.427 & 0.418 \\
    & $r_{\text{drop}}$ & 0.08 & 0.10 & 0.13 & 0.26 & 0.28 &0.30 & 0.65 & 0.72 & 0.49 & 0.52 & 0.57 \\
    \hline
    \multirow{2}{*}{Lego} 
    & $\overline{\sigma_{\text{max}}}$ & 0.344 & \textcolor{red}{0.354} & 0.325 & 0.363 & 0.352 &0.348 & 0.461 & 0.382 & 0.463 & 0.417 & 0.339 \\
    & $r_{\text{drop}}$ & 0.08 & 0.08 & 0.08 & 0.18 & 0.21 &0.23 & 0.47 & 0.50 & 0.36 & 0.40 & 0.46 \\
    \hline
    \multirow{2}{*}{Drums} 
    & $\overline{\sigma_{\text{max}}}$ & 0.369 & 0.358 & 0.335 & 0.328 & \textcolor{red}{0.336} &0.311 & 0.462 & 0.427 & 0.501 & 0.489 & 0.430 \\
    & $r_{\text{drop}}$ & 0.04 & 0.04 & 0.06 & 0.15 & 0.15 &0.16 & 0.62 & 0.73 & 0.46 & 0.46 & 0.57 \\
    \hline
    \multirow{2}{*}{Ficus} 
    & $\overline{\sigma_{\text{max}}}$ & 0.300 & 0.287 & 0.252 & 0.318 & 0.273 &0.245 & 0.397 & 0.388 & 0.347 & 0.328 & 0.275 \\
    & $r_{\text{drop}}$ & 0.08 & 0.12 & 0.13 & 0.25 & 0.29 &0.31 & 0.72 & 0.77 & 0.66 & 0.67 & 0.69 \\
    \hline
    \multirow{2}{*}{Hotdog} 
    & $\overline{\sigma_{\text{max}}}$ & 0.340 & 0.319 & \textcolor{red}{0.321} & 0.329 & 0.308 &0.298 & 0.463 & 0.384 & 0.473 & 0.445 & 0.376 \\
    & $r_{\text{drop}}$ & 0.10 & 0.12 & 0.12 & 0.26 & 0.29 &0.30 & 0.64 & 0.70 & 0.43 & 0.48 & 0.60 \\
    \bottomrule
    \end{tabular}%
}
\end{small}
\setlength{\tabcolsep}{6pt}
\caption{\textbf{Active Learning Scenario on Blender dataset}: PH-Dropout robustness to active learning is showed by a  decreasing epistemic uncertainty, $\overline{\sigma_{\text{max}}}$, at a similar dropout rate $r_{drop}$, or a stable $\overline{\sigma_{\text{max}}}$ at increasing $r_{drop}$, with increasing number of training views, given a constant $\epsilon$. The cases where PH-DROPOUT does not adhere to the active learning principle are marked with \textcolor{red}{red}.}
\label{table:active_bounded}
\end{table*}

\begin{table*}
\centering
\setlength{\tabcolsep}{4pt}
\begin{small}
\resizebox{0.5\columnwidth}{!}{%
    \begin{tabular}{l|c|c|c|c|c|c|c|c|c}
    \toprule
    Method & \multicolumn{5}{c}{2DGS} & \multicolumn{4}{c}{3DGS}\\
    \hline
    \multirow{2}{*}{Metric} & \multicolumn{8}{c}{Viewpoints (16v, 64v, 128v, 256v)} \\
    \cline{2-10} 
    & & 16v & 64v & 128v & 256v & 16v & 64v & 128v & 256v \\
    \hline
    \multirow{2}{*}{Train} 
    & $\overline{\sigma_{\text{max}}}$ & 0.390 & \textcolor{red}{0.397} & \textcolor{red}{0.408} & \textcolor{red}{0.409} & 0.338 & 0.328 & 0.321 & 0.317 \\
    & $r_{\text{drop}}$ & 0.16 & 0.26 & 0.32 & 0.38 & 0.12 & 0.15 & 0.18 & 0.20 \\
    \hline
    \multirow{2}{*}{Truck} 
    & $\overline{\sigma_{\text{max}}}$ & 0.435 & 0.424 & 0.413 & 0.407 & 0.369 & 0.342 & 0.331 & 0.329 \\
    & $r_{\text{drop}}$ & 0.21 & 0.32 & 0.36 & 0.40 & 0.14 & 0.18 & 0.21 & 0.23\\
    \bottomrule
    \end{tabular}%
}
\end{small}
\caption{\textbf{Unbounded Scenarios:} Performance of \method on quantifying epistemic uncertainty in GS-based methods}
\label{table:GS_al_unbounded}
\end{table*}

\subsection{Detailed correlation between RMSE Map and UQ with \method}\label{ap:rho}
Here we include the per scenario correlation between RMSE and UQ in the following tables, Table~\ref{table:fid_bounded} and Table~\ref{table:train_truck}.

\begin{table*}
\centering
\setlength{\tabcolsep}{4pt}
\begin{small}
\resizebox{\columnwidth}{!}{%
    \begin{tabular}{l|cc|cc|cc|cc|cc|cc|cc|cc|cc|cc|cc|cc}
    \toprule
    Method& \multicolumn{6}{c}{FreeNeRF}& \multicolumn{6}{c}{FreeNeRF+SPE} & \multicolumn{4}{c}{2DGS} & \multicolumn{6}{c}{3DGS}\\
    \hline
    \multirow{2}{*}{Dataset} & \multicolumn{2}{c}{8v}  & \multicolumn{2}{c}{16v}  & \multicolumn{2}{c}{100v}
    & \multicolumn{2}{c}{8v}  & \multicolumn{2}{c}{16v}  & \multicolumn{2}{c}{100v}
    & \multicolumn{2}{c}{16v}  & \multicolumn{2}{c}{100v}
    & \multicolumn{2}{c}{8v}  & \multicolumn{2}{c}{16v}  & \multicolumn{2}{c}{100v}\\
    \cline{2-25} 
    &$\rho_{\text{s}}$ $\uparrow$ &$\rho_{\text{p}}$ $\uparrow$ &$\rho_{\text{s}}$ $\uparrow$ &$\rho_{\text{p}}$  $\uparrow$ &$\rho_{\text{s}}$ $\uparrow$ & $\rho_{\text{p}}$ $\uparrow$
    &$\rho_{\text{s}}$ $\uparrow$ &$\rho_{\text{p}}$ $\uparrow$ &$\rho_{\text{s}}$ $\uparrow$ &$\rho_{\text{p}}$  $\uparrow$ &$\rho_{\text{s}}$ $\uparrow$ & $\rho_{\text{p}}$ $\uparrow$
    &$\rho_{\text{s}}$ $\uparrow$ &$\rho_{\text{p}}$  $\uparrow$ &$\rho_{\text{s}}$ $\uparrow$ & $\rho_{\text{p}}$ $\uparrow$
    &$\rho_{\text{s}}$ $\uparrow$ &$\rho_{\text{p}}$ $\uparrow$ &$\rho_{\text{s}}$ $\uparrow$ &$\rho_{\text{p}}$$\uparrow$ &$\rho_{\text{s}}$ $\uparrow$ & $\rho_{\text{p}}$ $\uparrow$
    \\
    \hline
      Mic  &0.981  &0.412 &0.985 &0.396 & 0.984 & 0.397 & 0.984 & 0.435 & 0.983 & 0.425 &0.982  &0.415 & 0.996 & 0.693 & 0.996 & 0.750 & 0.996 & 0.720 & 0.997 & 0.754 & 0.997 & 0.776\\
    \hline
    Chair & 0.963 & 0.388 & 0.962 & 0.338 & 0.967 & 0.407 & 0.967 & 0.405 & 0.967 & 0.400 &0.970 &0.412 & 0.990 & 0.739 & 0.993 & 0.817 & 0.989 & 0.682 & 0.991 & 0.742 & 0.994 & 0.830\\
    \hline
    Ship & 0.873 & 0.454 & 0.863 & 0.438 & 0.856 & 0.414 & 0.887 & 0.484 & 0.879 & 0.458 &0.891 & 0.486& 0.940 & 0.694 & 0.943 & 0.683 & 0.898 & 0.643 & 0.915 & 0.619 & 0.943 & 0.642\\
    \hline
    Materials & 0.918 & 0.329 & 0.934 & 0.361 & 0.868 & 0.164 & 0.939 & 0.372 & 0.943 & 0.400 &0.945 &0.410 & 0.979 & 0.680 & 0.982 & 0.695 & 0.964 & 0.669 & 0.982 & 0.669 & 0.984 & 0.678\\
    \hline
    Lego & 0.936 & 0.413 & 0.933 & 0.406 & 0.927 & 0.419  & 0.944 & 0.434 & 0.946 & 0.455 &0.949 &0.461 & 0.979 & 0.696 & 0.986 & 0.769 & 0.975 & 0.671 & 0.979 & 0.692 & 0.987 & 0.792\\
    \hline
    Drums & 0.961 & 0.472 & 0.947 & 0.378 & 0.953 & 0.392 & 0.963 & 0.497 & 0.959 & 0.453 &0.961 &0.474 & 0.980 & 0.649 & 0.990 & 0.703 & 0.980 & 0.660 & 0.988 & 0.652 & 0.991 & 0.673\\
    \hline
    Ficus & 0.978 & 0.479 & 0.984 & 0.547 & 0.982 &  0.547 & 0.982 & 0.527 & 0.986 & 0.573 &0.988 &0.582 & 0.992 & 0.718 & 0.993 & 0.782 & 0.994 & 0.752 & 0.995 & 0.779 & 0.996 & 0.844\\
    \hline
    Hotdog & 0.941 & 0.376 & 0.946 & 0.404 & 0.944 & 0.416 & 0.946 & 0.401 & 0.951 & 0.419 &0.953 &0.421 & 0.978 & 0.752 & 0.987 & 0.813 & 0.972 & 0.644 & 0.978 & 0.690 & 0.987 & 0.789\\
    \bottomrule
    \end{tabular}%
    }
    \end{small}
\setlength{\tabcolsep}{6pt}
\caption{\textbf{Bounded Scenarios}: Performance on Blender dataset with different training views. $\rho_{\text{s}}$: Spearman correlation, $\rho_{\text{p}}$: Pearson correlation.}
\label{table:fid_bounded}
\end{table*}


\begin{table*}[t]
\centering
\setlength{\tabcolsep}{4pt}
\begin{small}
\resizebox{\columnwidth}{!}{%
    \begin{tabular}{l|cc|cc|cc|cc|cc|cc|cc|cc}
    \toprule
    Method & \multicolumn{8}{c}{2DGS} & \multicolumn{8}{c}{3DGS}\\
    \hline
    \multirow{2}{*}{Dataset} & \multicolumn{2}{c}{16v}  & \multicolumn{2}{c}{64v}  & \multicolumn{2}{c}{128v}  & \multicolumn{2}{c}{256v} 
    & \multicolumn{2}{c}{16v}  & \multicolumn{2}{c}{64v}  & \multicolumn{2}{c}{128v}  & \multicolumn{2}{c}{256v} \\
    \cline{2-17} 
    & $\rho_{\text{s}}$ $\uparrow$ & $\rho_{\text{p}}$ $\uparrow$ & $\rho_{\text{s}}$ $\uparrow$ & $\rho_{\text{p}}$ $\uparrow$ 
    & $\rho_{\text{s}}$ $\uparrow$ & $\rho_{\text{p}}$ $\uparrow$ & $\rho_{\text{s}}$ $\uparrow$ & $\rho_{\text{p}}$ $\uparrow$
    & $\rho_{\text{s}}$ $\uparrow$ & $\rho_{\text{p}}$ $\uparrow$ & $\rho_{\text{s}}$ $\uparrow$ & $\rho_{\text{p}}$ $\uparrow$
    & $\rho_{\text{s}}$ $\uparrow$ & $\rho_{\text{p}}$ $\uparrow$ & $\rho_{\text{s}}$ $\uparrow$ & $\rho_{\text{p}}$ $\uparrow$ \\
    \hline
    Train & 0.270 & 0.213 & 0.351 & 0.308 & 0.354 & 0.317 & 0.388 & 0.355 & 0.299 & 0.239 & 0.362 & 0.328 & 0.384 & 0.345 & 0.412 & 0.360 \\
    \hline
    Truck & 0.311 & 0.293 & 0.414 & 0.429 & 0.432 & 0.439 & 0.426 & 0.436 & 0.343 & 0.333 & 0.371 & 0.419 & 0.393 & 0.420 & 0.410 & 0.420 \\
    \bottomrule
    \end{tabular}%
}
\end{small}
\setlength{\tabcolsep}{6pt}
\caption{\textbf{Unbounded Scenarios}: Performance of Gaussian Splatting methods on T\&T dataset with different training views. $\rho_{\text{s}}$: Spearman correlation, $\rho_{\text{p}}$: Pearson correlation.}
\label{table:train_truck}
\end{table*}

\subsection{2DGS in few-view cases}\label{ap:2dgs}
\begin{figure}[t]
\centering
  \includegraphics[width=0.8\textwidth]{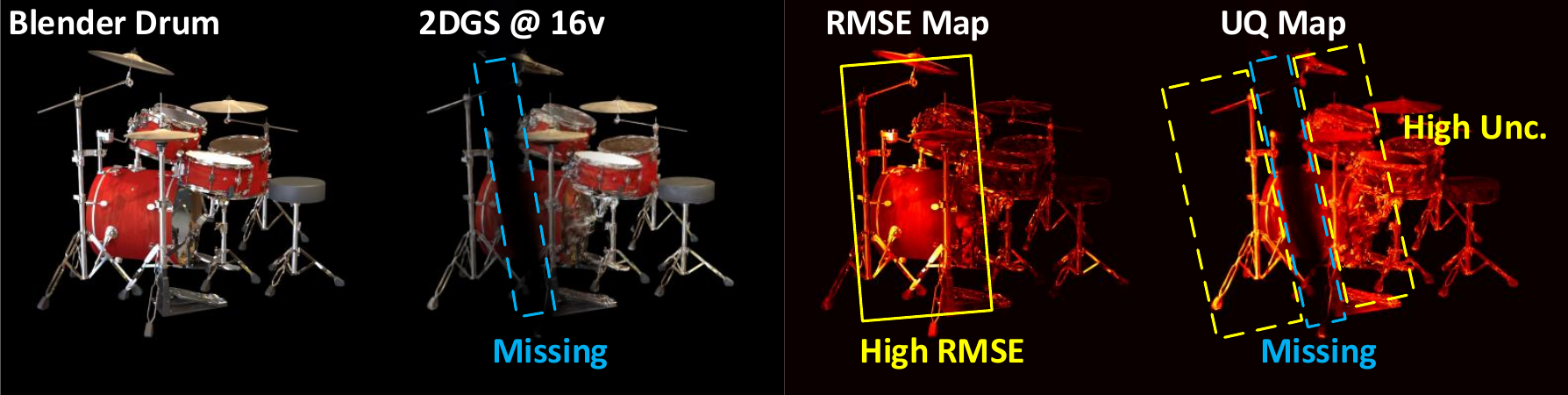}
\caption{2DGS misses rendering certain part of the object, when it is trained with few-view, \emph{e.g.}, 16-views for blender dataset. Besides the missing part, \method is able to show UQ with clear correlation with RMSE.}
  \label{fig:2dgs_miss}
\end{figure}

As Figure~\ref{fig:2dgs_miss} shows, 2DGS encounters a similar issue to hash-encoding-based NeRF, where certain parts of the object fail to render entirely, limiting \method's ability to detect significant variance. Without this variance, \method cannot effectively perform uncertainty quantification (UQ).

\subsection{Details of Ensemble Usecase}\label{ap:usecase}
Here we enclose more details about the ensemble usecase in \S\ref{sec:usecase}, including Table~\ref{table:uncertainty_drive_ensemble} and Table~\ref{table:uncertainty_drive_ensemble_unbound}.

\begin{table*}[t]
\centering
\setlength{\tabcolsep}{4pt}
\begin{small}
\resizebox{0.65\columnwidth}{!}{%
    \begin{tabular}{l|c|c|c|c|c|c|c|c|c|c}
    \toprule
    Method & \multicolumn{5}{c}{2DGS} & \multicolumn{5}{c}{3DGS}\\
    \hline
    Dataset 
    &Metric & 16v-a & 16v-b  &selected & $E_\mathrm{ME}$ & 16v-a & 16v-b &selected & $E_\mathrm{ME}$ \\
    \hline
    \multirow{2}{*}{Mic} 
    & SSIM & 0.921 & \underline{0.928} & \textbf{0.931} & 0.991 & \textbf{0.947} & 0.944 & \textbf{ 0.947} & 0.997 & \\
    & PSNR & \underline{24.7} & \textbf{25.1} & 24.5 & & \textbf{27.4} & 26.9 & \underline{27.2} & &\\
    \hline
    \multirow{2}{*}{Chair} 
    & SSIM & 0.920 & \underline{0.927} & \textbf{0.935} & 0.991 & 0.931 & \underline{0.935} & \textbf{0.938} & 0.992 &\\
    & PSNR & 25.6 & \underline{26.0} & \textbf{26.1} & & 26.5 & \underline{27.0} & \textbf{27.2} & &\\
    \hline
    \multirow{2}{*}{Ship} 
    & SSIM & \textbf{0.793} & 0.742 & \underline{0.747} & 0.912 & \underline{0.781} & 0.777 & \textbf{0.800} & 0.993 &\\
    & PSNR & \textbf{24.7} & 22.8 & \underline{23.1} & & \underline{25.5} & 25.2 & \textbf{26.1} & &\\
    \hline
    \multirow{2}{*}{Materials} 
    & SSIM & \textbf{0.871} & 0.857 & \underline{0.863} & 0.978 & \underline{0.892} & 0.884 & \textbf{0.903} & 1.00 &\\
    & PSNR & \textbf{23.3} & 21.7 & \underline{22.6} & & \underline{24.8} & 24.3 & \textbf{25.6} & &\\
    \hline
    \multirow{2}{*}{Lego} 
    & SSIM & \underline{0.907} & 0.902 & \textbf{0.910} & 0.995 & \underline{0.916} & 0.915 & \textbf{0.925} & 0.999 &\\
    & PSNR & \underline{26.8} & 25.6 & \textbf{27.0} & & \underline{28.0} & 27.6 & \textbf{28.3} & &\\
    \hline
    \multirow{2}{*}{Drums} 
    & SSIM & \textbf{0.859} & 0.777 & \underline{0.818} & 0.935 & \underline{0.890} & 0.851 & \textbf{0.901} & 0.998 &\\
    & PSNR & \textbf{19.9} & 18.9 & \underline{19.8} & & \underline{22.6} & 21.1 & \textbf{23.1} & &\\
    \hline
    \multirow{2}{*}{Ficus} 
    & SSIM & 0.917 & \textbf{0.933} & \underline{0.924} & 0.988 & 0.932 & \underline{0.935} & \textbf{0.936} & 0.998 &\\
    & PSNR & 23.7 & \textbf{25.7} & \underline{24.5} & & 25.6 & \underline{25.9} & \textbf{26.0} & &\\
    \hline
    \multirow{2}{*}{Hotdog} 
    & SSIM & 0.909 & \underline{0.920} & \textbf{0.923} & 0.998 & 0.926 & \underline{0.944} & \textbf{0.945} & 0.998 &\\
    & PSNR & 26.1 & \underline{26.6} & \textbf{27.1} & & 26.9 & \underline{29.5} & \textbf{29.7} & &\\
    \hline
    \multirow{2}{*}{Avg.} 
    & SSIM & \textbf{0.887} & 0.873 & \underline{0.881} & 0.974 & \underline{0.900} & 0.894 & \textbf{0.912} & 0.996 &\\
    & PSNR & \textbf{24.4} & 24.1 & \underline{24.3} & & \underline{25.4} & 25.3 & \textbf{26.3} & &\\
    \bottomrule
    \end{tabular}%
}
\end{small}
\vspace{-0.7em}
\caption{\textbf{Bounded Scenarios.} Select synthetic view from models with different training views, so that two models are merged on-the-fly.}
\label{table:uncertainty_drive_ensemble}
\end{table*}

\begin{table*}[t]
\centering
\setlength{\tabcolsep}{4pt}
\begin{small}
\resizebox{0.65\columnwidth}{!}{%
    \begin{tabular}{l|c|c|c|c|c|c|c|c|c}
    \toprule
    Method & \multicolumn{5}{c}{2DGS} & \multicolumn{4}{c}{3DGS}\\
    \hline
    Dataset 
    &Metric & 16v-a & 16v-b &selected & $E_\mathrm{ME}$ & 16v-a & 16v-b &selected & $E_\mathrm{ME}$ \\
    \hline
    \multirow{2}{*}{Train} 
    & SSIM & 0.476 & \underline{0.491} & \textbf{0.522} & 0.940 & 0.463 & \underline{0.468} & \textbf{0.526} & 0.979 \\
    & PSNR & 13.3 & \underline{14.5} & \textbf{14.9} & & 13.9 & \underline{14.5} & \textbf{15.5} \\
    \hline
    \multirow{2}{*}{Truck} 
    & SSIM & \underline{0.652} & 0.615 & \textbf{0.660} & 0.963 & \underline{0.636} & 0.606 & \textbf{0.676} & 0.995 \\
    & PSNR & \underline{17.5} & 16.7 & \textbf{17.7} & & \underline{17.7} & 17.0 & \textbf{18.6} \\
    \hline
    \multirow{2}{*}{Avg.} 
    & SSIM & \underline{0.564} & 0.553 & \textbf{0.591} & 0.952 & \underline{0.550} & 0.537 & \textbf{0.601} & 0.987 \\
    & PSNR & 15.4 & \underline{15.6} & \textbf{16.3} & & 15.8 & 15.8 & \textbf{17.1} \\
    \bottomrule
    \end{tabular}%
}
\end{small}
\vspace{-0.5em}
\caption{\textbf{Unbounded Scenarios.} Select synthetic view from models with different training views, so that two models are merged on-the-fly.}
\label{table:uncertainty_drive_ensemble_unbound}
\end{table*}

\end{document}